\documentclass[12pt,a4paper]{article}

\usepackage{float,amsmath,amsthm,mathtools,wrapfig,algorithm,algorithmicx}
\usepackage{algpseudocode}
\usepackage{graphicx}
\usepackage[font=small]{caption}
\usepackage{hyperref} 
\usepackage[nottoc]{tocbibind}
\usepackage[authoryear]{natbib}
\usepackage{centernot}

\usepackage[utf8]{inputenc}
\usepackage[charter]{mathdesign}

\allowdisplaybreaks[1]

\hypersetup{
    colorlinks=true,
    menucolor=black,
    citecolor=blue,
    linktocpage=true
}

\title{Implementing a Bayes Filter in a Neural Circuit: The Case of Unknown Stimulus Dynamics}
\author{Sacha Sokoloski \\ \small{Max Planck Institute for Mathematics in the Sciences}}
\date{}

\newcommand{\eprms}{{\V \theta}}
\newcommand{\iprms}{{\V \Theta}}

\newcommand\V[1]{\ensuremath{\boldsymbol{\mathbf{#1}}}}
\newcommand{\E}{{\mathbb E}}

\newcommand{\N}{{\mathbb N}}

\begin{document}

\maketitle

\begin{abstract}
    In order to interact intelligently with objects in the world, animals must first transform neural population responses into estimates of the dynamic, unknown stimuli which caused them. The Bayesian solution to this problem is known as a Bayes filter, which applies Bayes' rule to combine population responses with the predictions of an internal model. The internal model of the Bayes filter is based on the true stimulus dynamics, and in this paper we present a method for training a theoretical neural circuit to approximately implement a Bayes filter when the stimulus dynamics are unknown. To do this we use the inferential properties of linear probabilistic population codes to compute Bayes' rule, and train a neural network to compute approximate predictions by the method of maximum likelihood. In particular, we perform stochastic gradient descent on the negative log-likelihood of the neural network parameters with a novel approximation of the gradient. We demonstrate our methods on a finite-state, a linear, and a nonlinear filtering problem, and show how the hidden layer of the neural network develops tuning curves which are consistent with findings in experimental neuroscience.
\end{abstract}

\section{Introduction}

Whether its concerns the location of distant food or the presence of a lurking predator, animals must reason about the world based on uncertain beliefs. The Bayesian brain is the hypothesis that the brain represents these beliefs with probability distributions, and reasons about the world based on the principles of Bayesian inference \citep{knill_bayesian_2004,doya_bayesian_2007}. The Bayesian brain is supported by both theoretical arguments \citep{jaynes_probability_2003} and experimental evidence \citep{ernst_humans_2002,fischer_owls_2011,fetsch_neural_2011,coen-cagli_flexible_2015}, yet many open questions remain in how exactly the brain implements Bayesian inference in populations of neurons \citep{bowers_bayesian_2012, pouget_probabilistic_2013}.

The goal of this work is to understand how neural circuits compute accurate beliefs about dynamic stimuli with Bayesian inference. To do this we first assume that the neural populations in a given circuit encode probability distributions in their activity with probabilistic population codes \citep{zemel_probabilistic_1998}. A key property of probabilistic population codes for Bayesian theories of the brain is that theoretical neural circuits based on probabilistic population codes can trivially implement Bayes' rule \citep{ma_bayesian_2006}.

Bayes' rule is the most fundamental equation in Bayesian inference, and describes how to compute optimal beliefs about an unknown stimulus by combining prior beliefs with observations of the stimulus. Nevertheless, Bayes' rule alone can only be applied to independent observations of static stimuli, and is not sufficient for explaining how animals compute beliefs about dynamic stimuli. The dynamic extension of Bayesian inference is known as Bayesian filtering, which complements Bayes' rule with an equation for computing optimal predictions \citep{thrun_probabilistic_2005,sarkka_bayesian_2013}. By combining prediction and online inference, it is possible to define the Bayes filter, which is an algorithm for computing optimal beliefs about unknown, dynamic stimuli.

The Kalman filter is an example of a Bayes filter for when the stimulus dynamics are known and linear. \cite{beck_marginalization_2011} show how to implement a form of Kalman filter in a theoretical neural circuit by combining the inferential properties of linear probabilistic population codes with a recurrent neural network which computes predictions. In order to define the neural network, they derive a closed-form expression for the network parameters based on the parameters of the linear stimulus dynamics. However, animals do not generally have built-in knowledge of the dynamics of stimuli, nor can they assume that these dynamics are linear. As such, this work is not sufficient for explaining how animals maintain accurate beliefs about dynamic stimuli.

In this paper we generalize the approach of \cite{beck_marginalization_2011} to arbitrary dynamical systems where the stimulus dynamics are unknown. To do this, we begin by replacing the derived, linear network of \cite{beck_marginalization_2011} with a general network with tunable parameters. By taking advantage of the exponential family structure of linear probabilistic population codes \citep{welling_exponential_2004,beck_probabilistic_2007}, we then show how to minimize the negative log-likelihood of the parameters of the network with stochastic gradient descent. Finally, we develop a novel algorithm for approximating the true stochastic gradient, which we compare against contrastive divergence minimization \citep{hinton_training_2002}. Although in our demonstrations we define the recurrent neural network as a form of multilayer perceptron, the theory we present can be applied to any parameterized network architecture, in particular ones which satisfy more realistic biological constraints.

The theory we develop in this paper is related to two additional approaches in the machine learning and computational neuroscience literatures. Firstly, the theoretical neural circuit that we construct can be roughly interpreted as a form of the model presented in \cite{boulanger-lewandowski_modeling_2012} for approximate filtering. In our approach, however, the neural circuit satisfies a set of equations which ensure that Bayes' rule is exactly implemented, which is not present in the work of \cite{boulanger-lewandowski_modeling_2012}. Secondly, \cite{makin_learning_2015} present an alternative approach to implementing a Bayes filter in a neural circuit based on probabilistic population codes. The form of the circuit in \cite{makin_learning_2015}, however, is markedly different from our own, and although it lacks some of the theoretical advantages of our neural circuit, it makes use of a more biologically realistic learning rule. We discuss these related methods in more detail at the end of this paper.

For the purposes of demonstration we apply our methods in three simulated experiments, each of which models how a particular neural circuit learns to maintain accurate beliefs about some unknown stimulus. In the first simulation the stimulus is a set of colours in a sequence learning task, which we model with a 3-state Markov chain, and the neural circuit is composed of colour recognition and sequence learning neural populations along the ventral stream. In the second simulation the stimulus is the position of a mouse on a track, which we model as a linear dynamical system, and the neural circuit is part of the self-localization system of the hippocampus. In the third simulation the stimulus is the angular position and velocity of a human arm, which we model as a stochastic pendulum, and the neural circuit is part of the proprioceptive system in the cerebellum.

In the first and second simulations it is possible to compute the optimal beliefs of the Bayes filter based on the true stimulus dynamics, whereas in the third simulation, approximate beliefs can be computed with a form of extended Kalman filter. These filters provide ground truth for our theoretical neural circuits, and in all cases, we find that our circuits are able to maintain good approximate beliefs about the stimulus. Moreover, by analyzing the hidden layer of the multilayer perceptron in the third experiment, we show how the network uses gain-fields to represent position and velocity information in a manner that is consistent with theory and experiment \citep{sejnowski_spatial_1995, salinas_gain_2000, paninski_superlinear_2004, herzfeld_encoding_2015}. At the end of this paper we consider additional features of our work which are relevant for neuroscience, and present ways in it can be extended in the future.

\section{Inference}
\label{sec:inference}

Statistical inference is the process of estimating unknown quantities through observation. There are many paradigms for formalizing statistical inference, but Bayesian inference is arguably the most appropriate framework for describing how an agent maintains subjective beliefs which correspond to the world. Two of the most well-known theoretical arguments for Bayesian inference are Cox's theorem and the class of Dutch book arguments \citep{jaynes_probability_2003,talbott_bayesian_2015}. On one hand, Cox's theorem demonstrates that the consistent extension of propositional logic on binary truth values to continuous probabilities is given by Bayesian inference. On the other hand, Dutch book arguments demonstrate that failing to follow the principles of Bayesian inference can lead gamblers to make wagers which they are guaranteed to lose.

Both of these arguments begin with the assumption that subjective beliefs are represented by probabilities, and lead irrevocably to Bayesian inference. In the context of neuroscience, this implies that if populations of neurons represent beliefs with encoded probability distributions, then populations of neurons must extract information from observations in accordance with the principles of Bayesian inference. Nevertheless, how populations of neurons implement Bayesian inference remains an open question, because there are many proposals for how populations of neurons might encode probability distributions \citep{pouget_probabilistic_2013}.

Each coding scheme has its own advantages, and greatly simplify particular operations on encoded information. Linear probabilistic population codes (LPPCs) have the invaluable property that they allow Bayes' rule to be trivially implemented on encoded beliefs \citep{ma_bayesian_2006}. Evaluating Bayes' rule is the most fundamental operation in Bayesian inference, and involves combining the likelihood of an observation with prior beliefs in order to compute posterior beliefs. By using LPPCs, \cite{ma_bayesian_2006} demonstrate how to construct a neural circuit which computes encodings of posterior beliefs by taking a linear combination of the response of a neural population and encoded prior beliefs.

A concrete example of such a neural circuit is the self-localization system in the hippocampus and entorhinal cortex \citep{moser_place_2008}. Place cells in the hippocampus encode information about the position of the animal, and many studies suggest that place cells optimize this encoding through a combination of local activity in the limbic system and incoming sensory information \citep{mcnaughton_path_2006}. In the Bayesian picture, place cells encode posterior beliefs about the position of the animal, and update these beliefs with Bayes' rule given the responses of sensory neurons.

We begin this section by introducing Bayes' rule, and we then formally define populations of neurons which generate Poisson-distributed spikes in response to stimuli. We then formally introduce LPPCs, and demonstrate their relationship with the family of machine learning models known as exponential family harmoniums \citep{smolensky_information_1986,welling_exponential_2004}. This relationship allows us to easily rederive the results of \cite{ma_bayesian_2006}, and later allows us to apply machine learning algorithms to gather learning statistics from LPPCs and thereby model how neural systems learn to solve the filtering problem.

\subsection{Bayes' Rule}
\label{sec:bayes-rule}

Assuming it exists, the joint density $p_{XN}$ provides a complete description of the relationship between the pair of random variables $X$ and $N$. Given the joint density $p_{XN}$, we may derive the marginal densities $p_X$ and $p_N$, and the conditional densities $p_{X \mid N}$ and $p_{N \mid X}$. With a bit of algebra, we may also derive Bayes' rule
\begin{equation}
    p_{X \mid N}(\V x \mid \V n) = \frac{p_{N \mid X}(\V n \mid \V x)p_X(\V x)}{p_N(\V n)}.
\end{equation}

Bayes' rule itself is a simple equation, but its interpretation is the basis for a general approach to statistical inference. Since we are applying Bayesian inference in the context of neuroscience, let us refer to $\V n$ as the response and $\V x$ as the stimulus. In Bayesian inference, $p_X$ is the prior, which represents our current beliefs about the stimulus, and $p_{N \mid X}$ is the likelihood, which describes how responses are generated to the stimulus. The density $p_{X \mid N}$ is the posterior, which represents our new beliefs about the stimulus after observing a response, and all of these densities may be derived from the generative model $p_{XN}$, which is a complete description of the probabilistic relationship between the stimulus and response.

For a given response $\V n$, $p_N(\V n)$ is constant. Since the posterior $p_{X \mid N = \V n}$ is a probability density and must therefore integrate to 1, knowing the posterior up to a constant factor is sufficient for determining the posterior. For this reason, Bayes' rule is often defined as the proportionality relation
\begin{equation}
    p_{X \mid N}(\V x \mid \V n) \propto p_{N \mid X}(\V n \mid \V x)p_X(\V x),
    \label{eq:bayes-rule}
\end{equation}
which emphasizes that knowing the prior and the likelihood is sufficient for determining the posterior.

\subsection{Poisson Neurons}
\label{sec:poisson-neurons}

Poisson neurons are a simple, yet theoretically rigorous approach to modelling how neurons respond to stimuli \citep{dayan_theoretical_2005}, and we make use of them throughout this paper. Given a population of $d_N$ Poisson neurons and a stimulus $\V x$, the $i$th Poisson neuron generates a Poisson-distributed number of spikes with rate $\gamma f_i(\V x)$, where $f_i(\V x)$ is the tuning curve of the neuron, and $\gamma$ is the gain. The component neurons of the population are conditionally independent of each other given the stimulus, such that the likelihood with respect to the entire population response $N$ may be written
\begin{equation}
    p_{N \mid X}(\V n \mid \V x) = \prod_{i=1}^{d_N} p_{N_i \mid X}(n_i \mid \V x) = \prod_{i = 1}^{d_N} \frac{e^{-\gamma f_i(\V x)} (\gamma f_i(\V x))^{n_i}}{n_i!}.
    \label{eq:poisson-likelihood}
\end{equation}

\begin{figure}[t!]
    \centering
    \includegraphics[width=\textwidth]{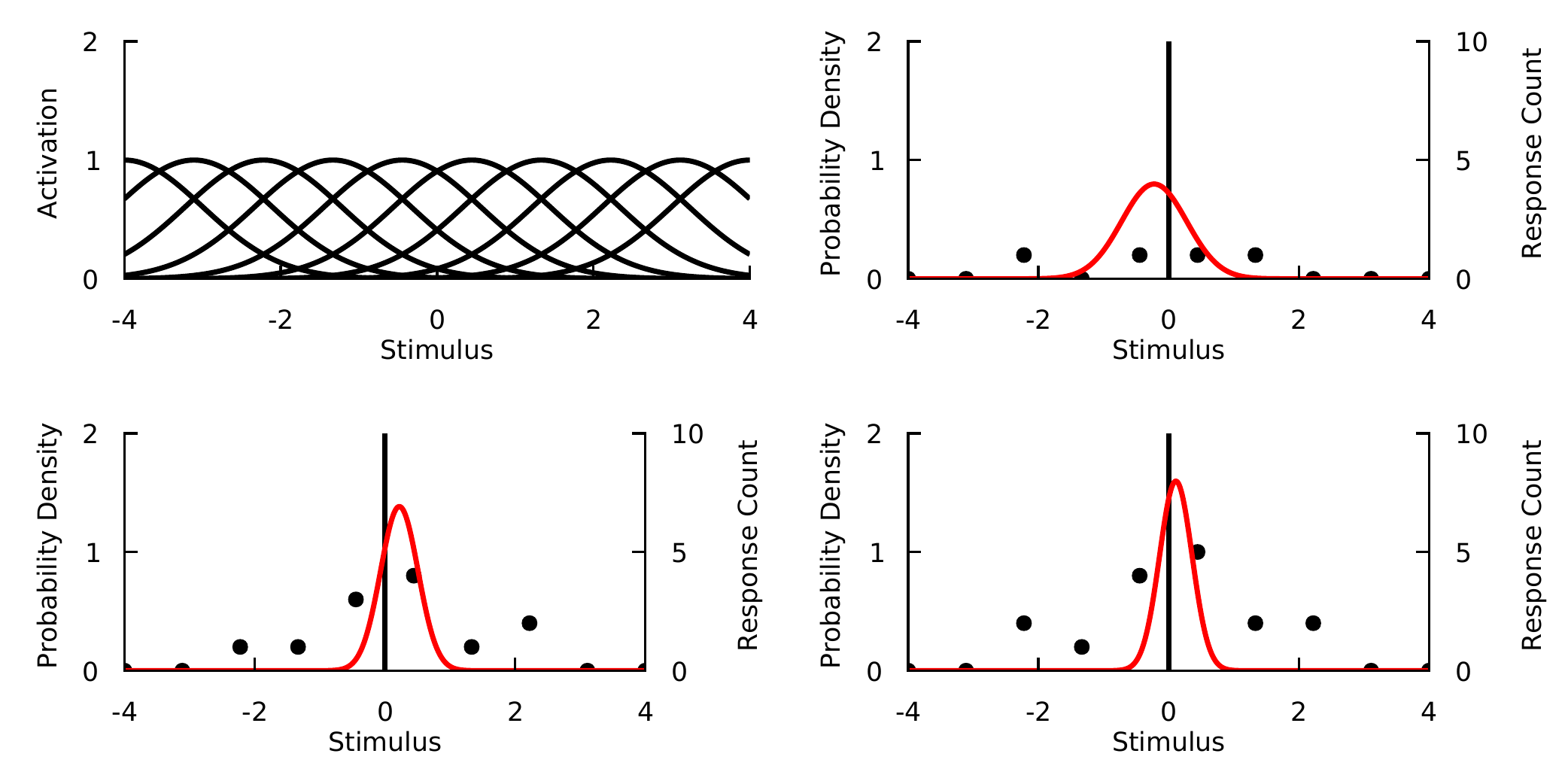}
    \caption{\textbf{Linear Probabilistic Population Codes}: These plots demonstrate encoding and decoding with linear probabilistic population codes. \emph{Top Left}: The rates of ten Gaussian tuning curves with uniformly distributed preferred stimuli, and gain $\gamma = 1$. \emph{Top Right}: The components of the response $\V n_1$ (black dots) to the stimulus 0 (black line) generated with $\gamma = 2$, and the resulting posterior $p_{X \mid N = \V n_1}$ (red line) based on an approximately flat prior. \emph{Bottom Left}: Response $\V n_2$, generated with $\gamma = 4$. \emph{Bottom Right}: The encoded posterior of the rate of the posterior population $\V z = \V n_2 + \V y$, where $\V A = \V B = \V I$, and the rate of the prior population is $\V y = \V n_1$. Note that $\V z$ is a more accurate encoding of the stimulus than either $\V y$ or $\V n_2$ alone.}
    \label{fig:population-code}
\end{figure}

When example tuning curves are called for in this paper, we consider the 1-dimensional Gaussian tuning curve
\begin{equation}
    f_i(x) = e^{\frac{-(x - x^0_i)^2}{2\sigma^2}},
    \label{eq:gaussian-tuning-curve}
\end{equation}
with preferred stimuli $x^0_i$ (figure \ref{fig:population-code}: Top Left). The theory we develop may nevertheless be generalized to a variety of tuning curves, as we later demonstrate in our simulations.

\subsection{Linear Probabilistic Population Codes}
\label{sec:probabilistic-population-codes}

A population code is a description of how to encode information about a stimulus in the combined activity of a population of neurons, as well as how to decode this information from the population. For a random stimulus $X$ and random response $N$, a probabilistic population code (PPC) stochastically encodes information in neural populations by sampling from the likelihood $p_{N \mid X}$, and decodes this information by computing the posterior $p_{X \mid N}$ \citep{zemel_probabilistic_1998,beck_probabilistic_2007}. Finally, a linear probabilistic population code (LPPC) is a PPC where both the likelihood and posterior may be expressed as log-linear functions \citep{ma_bayesian_2006,pouget_probabilistic_2013}. In this section we formally define LPPCs by using the theory of exponential families.

An exponential family is a set of probability densities with a specific form \citep{amari_methods_2007,wainwright_graphical_2008,nielsen_statistical_2009}. Each exponential family $\mathcal M$ is defined by a sufficient statistic $\V s$ and a base measure $\nu$. In turn, each density $q$ in the exponential family $\mathcal M$ is given by
\begin{equation}
    q(\V x) \propto e^{\eprms \cdot \V s(\V x)}\nu(\V x),
    \label{eq:exponential-family}
\end{equation}
where $\eprms$ are the natural parameters which specify the particular density. A wide variety of families of probably densities such as the normal, von Mises, and categorical families are in fact exponential families, and working with exponential families is typically much easier than generic families of densities.

Recall that at every stimulus $\V x$, the likelihood $p_{N \mid X = \V x}$ of a population of Poisson neurons (\ref{eq:poisson-likelihood}) is a product of Poisson densities. The set of all products of Poisson densities is an exponential family with sufficient statistic equal to the identity function and base measure $\nu(\V n) = (n_1! \cdots n_{d_N}!)^{-1}$. As such, where $\mathcal M_N$ is the exponential family of products of Poisson densities, the likelihood $p_{N \mid X = \V x} \in \mathcal M_N$ for any $\V x$.

\begin{figure}
    \centering
    \includegraphics[scale=0.45]{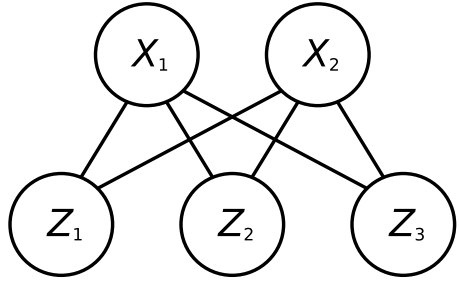}
    \caption{\textbf{Exponential Family Harmoniums}: Here we depict the graphical representation of an exponential family harmonium. From this graph we may infer that the component random responses $N_{(1)}$, $N_{(2)}$, and $N_{(3)}$ are mutually independent given the component random stimuli $X_{(1)}$ and $X_{(2)}$, and that $X_{(1)}$ is independent of $X_{(2)}$ given $N_{(1)}$, $N_{(2)}$, and $N_{(3)}$.}
    \label{fig:harmonium}
\end{figure}

Suppose that we also wish for the posterior $p_{X \mid N = \V n}$ at any response $\V n$ to be an element of some exponential family $\mathcal M_X$, where $\mathcal M_X$ is defined by the sufficient statistic $\V s$, and a base measure which, for simplicity, we assume to be constant. Then it can be shown \citep{besag_spatial_1974,arnold_compatible_1989, arnold_conditionally_2001} that the only generative model $p_{XN}$ consistent with the likelihood $p_{N \mid X}$ of a Poisson population and a posterior $p_{X \mid N}$ which satisfies these assumptions is another exponential family density with the log-linear form
\begin{equation}
    p_{XN}(\V x, \V n) \propto \frac{e^{\V s(\V x) \cdot \iprms_N \cdot \V n + \V s(\V x) \cdot \eprms_X + \V n \cdot \eprms_N}}{n_1! \cdots n_{d_N}!}.
    \label{eq:harmonium}
\end{equation}
In the machine learning literature, models with this form are known as exponential family harmoniums (\cite{welling_exponential_2004}, figure \ref{fig:harmonium}).

It follows from the theory of exponential family harmoniums \citep{welling_exponential_2004} that the posterior $p_{X \mid N}$ and likelihood $p_{N \mid X}$ of a generative model defined in this way may be expressed in the log-linear forms
\begin{equation}
    p_{N \mid X}(\V n \mid \V x) \propto \frac{e^{\V s(\V x) \cdot \iprms_N \cdot \V n + \V n \cdot \eprms_N}}{n_1! \cdots n_{d_N}!},
    \label{eq:likelihood}
\end{equation}
and
\begin{equation}
    p_{X \mid N}(\V x \mid \V n) \propto e^{\V s(\V x) \cdot \iprms_N \cdot \V n + \V s(\V x) \cdot \eprms_X},
    \label{eq:posterior}
\end{equation}
such that the likelihood $p_{N \mid X = \V x} \in \mathcal M_N$ has natural parameters $\V s(\V x) \cdot \iprms_N + \eprms_N$, and the posterior $p_{X \mid N = \V n} \in \mathcal M_X$ has natural parameters $\iprms_N \cdot \V n + \eprms_X$. Since a linear probabilistic population code is a probabilistic population code where the likelihood and posterior have log-linear forms, the conditional distributions of an exponential family harmonium define an LPPC.

Although it is perhaps surprising that the likelihood of a Poisson population (\ref{eq:poisson-likelihood}) can be expressed in the log-linear form of relation \ref{eq:likelihood}, the results of \cite{arnold_conditionally_2001} imply that this must be the case. For example, if $p_{N \mid X}$ is defined by the set of Gaussian tuning curves $f_i(x)$ with preferred stimuli $x^0_i$ and shared tuning width $\sigma$, then we can express it in the form of relation \ref{eq:likelihood} by setting the elements of the matrix $\iprms_N$ equal to
\begin{equation}
    \V \Theta_{N,(1,i)} = \frac{x^0_i}{\sigma^2},
    \hspace{8pt} \V \Theta_{N,(2,i)} = -\frac{1}{2\sigma^2},
\end{equation}
setting the elements of the vector $\eprms_N$ equal to
\begin{equation}
    \V \theta_{N,(i)} = \log \gamma - \frac{(x^0_i)^2}{2\sigma^2},
\end{equation}
and by letting $\V s(x) = (x,x^2)$, which is the sufficient statistic of the family of normal distributions (figure \ref{fig:population-code}: Top Right, Bottom Left).

\subsection{Stimulus-Independent Total Rate}
\label{sec:stimulus-independent-total-rate}

In many applications of Poisson neurons, the total rate $\gamma \sum_{i=1}^{d_N} f_i(X)$ of the population is designed to be approximately independent of the stimulus $X$ itself \citep{sejnowski_spatial_1995,pouget_inference_2003,beck_marginalization_2011}. For the total rate to be independent of the stimulus, the sum of the tuning curves must satisfy
\begin{equation}
    \sum_{i=1}^{d_N} f_i(\V x) = \lambda
    \label{eq:tuning-curve-sum}
\end{equation}
for some constant $\lambda$. It turns out that for many families of tuning curves, satisfying this relation is relatively straightforward. In the case of Gaussian tuning curves, for example, if one distributes the preferred stimuli uniformly over the space of the stimulus, then the sum of the tuning curves converges to a constant as the number of preferred stimuli is increased \citep{ma_bayesian_2006}.

In addition to providing expressions for the LPPC prior and posterior, it also follows from the theory of exponential family harmoniums that the LPPC prior may be expressed as
\begin{equation}
    p_X(\V x) \propto e^{\V s(\V x) \cdot \eprms_X + \gamma \sum_{i = 1}^{d_N} f_i(\V x)}.
    \label{eq:prior}
\end{equation}
Observe that when equation \ref{eq:tuning-curve-sum} is satisfied, the sum of the tuning curves in the LPPC prior is constant and can be absorbed into the constant of proportionality. This forces the LPPC prior to be an element of the same exponential family as the LPPC posterior. A prior is known as a conjugate prior when the prior and posterior share the same form. Conjugate priors have numerous computational advantages, and we return to equation \ref{eq:tuning-curve-sum} throughout this paper. One more advantage of LPPCs which satisfy this equation is that by setting $\eprms_X = \V 0$, we define the prior over the stimulus to be flat, which is a practical prior for many applications of Bayesian inference.

\subsection{Neural Bayes' Rule}
\label{sec:neural-bayes-rule}

Let us refer to the population of Poisson neurons which generate the responses $N$ as the observation population. In order to implement Bayesian inference in a theoretical neural circuit, let us define two further neural populations, called the prior population and the posterior population, with firing rates given by the random variables $Y$ and $Z$, and number of neurons equal to $d_Y$ and $d_Z$, respectively. Our goal is to define $Y$ and $Z$ such that they encode the prior and posterior densities in an application of Bayes' rule.

Now theoretically, by defining the random variables $Y$ and $Z$, we imply the existence of a generative model $p_{NXYZ}$, as well as implicit decoding densities $p_{X \mid Y}$ and $p_{X \mid Z}$. However, there are two reasons why we avoid relying on the implicit probabilistic structure of the random variables in order to define our decoders. Firstly, we ultimately know the form that we wish for the decoders to have, and ensuring that $p_{X \mid Y}$ and $p_{X \mid Z}$ have the desired forms requires fairly careful and rather abstruse analysis. Secondly, when it comes to the task of learning to approximate a Bayes filter, which we undertake in section \ref{sec:learning}, the implicit decoders are anyway inaccessible.

Therefore, in this section, and throughout this paper, we instead consider the prior and posterior decoders $q_{X \mid Y}$ and $q_{X \mid Z}$. We define these decoders to have forms of our choosing, and then demonstrate how to define $Y$ and $Z$ to encode the relevant densities in terms of these decoders. In those cases where we can perform exact inference, then these decoders are equal to the implicit decoders of the generative model. When we cannot perform exact inference, then we may still use these decoders to implement good approximations.

Let us define the density encoded by the prior rate vector $\V y$ as
\begin{equation}
    q_{X \mid Y}(\V x \mid \V y) \propto e^{\V s(\V x) \cdot \iprms_Y \cdot \V y + \gamma \sum_{i = 1}^{d_N} f_i(\V x)},
    \label{eq:encoded-prior}
\end{equation}
where $\iprms_Y$ is a matrix which we refer to as the decoding matrix of the prior population. Observe that $q_{X \mid Y = \V y}$ is equal to the prior in relation \ref{eq:prior} where $\eprms_X = \iprms_Y \cdot \V y$. Therefore, we may ensure that $q_{X \mid Y = \V y} = p_X$ by choosing rates of the prior population $\V y$ which satisfy $\eprms_X = \iprms_Y \cdot \V y$.

Let us then define the density encoded by the posterior rate vector $\V z$ as
\begin{equation}
    q_{X \mid Z}(\V x \mid \V z) \propto e^{\V s(\V x) \cdot \iprms_Z \cdot \V z},
    \label{eq:encoded-posterior}
\end{equation}
where $\iprms_Z$ is the decoding matrix of the posterior population. Let us also assume that $\iprms_Z$ is related to the matrix $\iprms_N$ of the likelihood by
\begin{equation}
    \iprms_N = \iprms_Z \cdot \V A,
    \label{eq:population-relation}
\end{equation}
for some matrix $\V A$, and related to the decoding matrix $\iprms_Y$ of the prior population by
\begin{equation}
    \iprms_Y = \iprms_Z \cdot \V B,
    \label{eq:population-relation2}
\end{equation}
for some matrix $\V B$.

Given these definitions, if we consider the LPPC posterior (\ref{eq:posterior}) given the prior rate vector $\V y$ and the response $\V n$, we find that
\begin{equation*}
    p_{X \mid N}(\V x \mid \V n) \propto e^{\V s(\V x) \cdot \iprms_N \cdot \V n + \V s(\V x) \cdot \iprms_Y \cdot \V y} = e^{\V s(\V x) \cdot \iprms_Z \cdot (\V A \cdot \V n + \V B \cdot \V y)},
\end{equation*}
which implies that
\begin{equation}
    p_{X \mid N}(\V x \mid \V n) = q_{X \mid Z}(\V x \mid \V A \cdot \V n + \V B \cdot \V y).
    \label{eq:neural-bayes-rule}
\end{equation}
We refer to this equation as neural Bayes' rule. In words, as long as the prior population encodes the true prior, and the observation, prior, and posterior populations satisfy equations \ref{eq:population-relation} and \ref{eq:population-relation2}, then Bayes' rule may be implemented as a linear combination of the rates of the prior population $\V y$ and the response of the observation population $\V n$ \citep{ma_bayesian_2006}.

\begin{figure}
    \centering
    \includegraphics[scale=0.45]{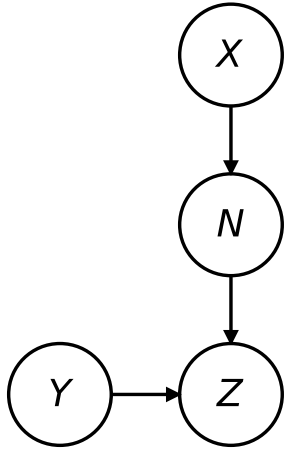}
    \caption{\textbf{Neural Circuit for Bayesian Inference}: Here we depict a random response $N$ from a population of Poisson neurons to a random stimulus $X$. The random rates $Y$ and $Z$ represent the rates of the prior and posterior populations, respectively. The random response $N$ is combined with the random prior rate $Y$ to compute the random posterior rate $Z$ in accordance with neural Bayes' rule.}
    \label{fig:neural-bayesian-inference}
\end{figure}

In accordance with neural Bayes' rule, if we define the rates of the posterior population by
\begin{equation}
    Z = \V A \cdot N + \V B \cdot Y,
    \label{eq:encoded-neural-bayes-rule}
\end{equation}
where $Y$ is drawn from a distribution over the set of $\V y$ that satisfy $\eprms_X = \iprms_Y \cdot \V y$, we may ensure that the posterior population encodes the posterior density for any realization of the circuit. We depict the neural circuit composed of $N$, $X$, $Y$, and $Z$ in figure \ref{fig:neural-bayesian-inference}, and we demonstrate an application of this circuit in the bottom right panel of figure \ref{fig:population-code}.

\section{Dynamics}
\label{sec:dynamics}

Bayesian inference formalizes statistical inference as the answer to the question, "What is the conditional probability distribution over the unknown variable given the available observations?" If we are given a sequence of population responses to a dynamic stimulus, we may compute the corresponding conditional probability over the stimulus with a recursive, two-step algorithm known as a Bayes filter. A Bayes filter recursively computes beliefs about the stimulus at time $k+1$ by computing predictions of the stimulus at time $k+1$ as a function of the beliefs at time $k$, and combining these predictions with the response at time $k+1$ using Bayes' rule \citep{thrun_probabilistic_2005,sarkka_bayesian_2013}. This has important implications for the Bayesian brain hypothesis, because if neural populations are performing Bayesian inference, then not only must populations of neurons implement Bayes' rule, but they must also implement prediction.

For some neuroscientists this is not surprising, as many argue that prediction is essential to neural computation and constitutes a guiding principle for the architecture of the brain \citep{bubic_prediction_2010,friston_free-energy_2010,clark_whatever_2013}. For example, forward models are ubiquitous in computational models of motor control \citep{miall_forward_1996,todorov_optimal_2002,franklin_computational_2011}. Forward models in the brain predict the result of a given motor command, and this information is combined with the responses of proprioceptors in order to estimate body position. In addition to the observation, prior, and posterior populations introduced in section \ref{sec:inference}, this entails the existence of an additional neural network which transforms efference copies of the motor command into predictions of the consequences of that command.

We begin this section by formalizing dynamic stimuli and dynamic Poisson populations, and continue by introducing the prediction and update equations which define the Bayes filter. We then introduce closed-form solutions to the Bayes filter on population responses for the case of finite-state systems and linear dynamical systems, generalizing an approach described in \cite{makin_learning_2015}. We conclude by demonstrating how these solutions can be implemented in theoretical neural circuits based on the work of \cite{beck_exact_2007} and \cite{beck_marginalization_2011}, and thereby lay the groundwork for how approximate solutions can be similarly defined.

\subsection{Dynamic Poisson Populations}

We define a dynamic Poisson population as a sequence of pairs of random variables $(X_k, N_k)_{k \in \N}$, where $X_k$ is the random stimulus at time $k$, and $N_k$ is the random response at time $k$. We assume that the dynamic stimulus is Markov, such that each stimulus $X_{k+1}$ is conditionally independent of the past random variables given the previous stimulus $X_k$, where $X_0$ is drawn from some initial density $p_{X_0}$. We also assume that the densities $p_{X_{k+1} \mid X_k}$ are invariant with respect to $k$, such that we may recursively generate the sample sequences of $(X_k)_{k \in \N}$ with a single, time-invariant conditional density, which we refer to as the transition density. Because the transition density is time-invariant, we denote it by $p_{X' \mid X} = p_{X_{k+1} \mid X_k}$ for any $k$.

\begin{figure}
    \centering
        \includegraphics[scale=0.45]{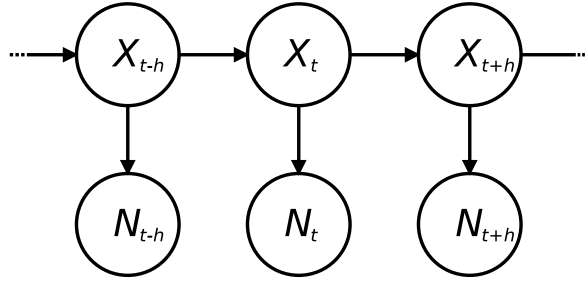}
        \caption{\textbf{Dynamic Poisson Population}: The graphical representation of  a dynamic Poisson population. The arrow from the current stimulus $X_k$ to the subsequent stimulus $X_{k+1}$ represents the dependence of the subsequent stimulus on the current stimulus, and is described by transition density $p_{X' \mid X}$. The arrow from the current stimulus $X_k$ to the current response $N_k$ represents the dependence of the response on the stimulus, and is described by emission density $p_{N \mid X}$.}
    \label{fig:hidden-markov-model}
\end{figure}

We assume that each response $N_k$ is conditionally independent of all the other random variables given the simultaneous response $X_k$, and that the conditional density $p_{N_k \mid X_k}$ is given by the likelihood of a population of Poisson neurons, as defined in equation \ref{eq:poisson-likelihood} and relation \ref{eq:likelihood}. The tuning curves and gain of the dynamic Poisson population often depend on time, but for the purposes of developing the theory in this section we assume that they are constant. As such, we denote the time-invariant conditional density by $p_{N \mid X} = p_{N_k \mid X_k}$ for any $k$, and refer to it as the emission density. We depict the graphical representation of the stimulus and Poisson population in figure \ref{fig:hidden-markov-model}.

\subsection{Bayesian Filtering}
\label{sec:bayesian-filtering}

Given the sequence of responses $\V n_0, \dots, \V n_k$ from a dynamic Poisson population, a Bayes filter is an algorithm for computing the beliefs $p_{X_k \mid N_0 = \V n_0, \dots, N_k = \V n_k}$. A Bayes filter is defined by two equations. The first is a form of Bayes' rule (\ref{eq:bayes-rule}), and is given by
\begin{align}
    \nonumber p_{X_{k+1} \mid N_0, \dots, N_{k+1}}&(\V x_{k+1} \mid \V n_0, \dots, \V n_{k+1}) \propto \\
        &p_{N \mid X}(\V n_{k+1} \mid \V x_{k+1})p_{X_{k+1} \mid N_0, \dots, N_k}(\V x_{k+1} \mid \V n_0, \dots, \V n_k).
    \label{eq:update-equation}
\end{align}
When normalized, this relation is known as the update equation, and computes the beliefs at time $k+1$ by applying Bayes' rule to the emission density $p_{N \mid X}$ and the prior $p_{X_{k+1} \mid N_0 = \V n_0, \dots, N_k = \V n_k}$.

This prior constitutes beliefs about the stimulus at time $k+1$ given only the sequence of responses up to time $k$. Because these prior beliefs transform available information into information about the future, these prior beliefs are known as predictions. The prediction density may be expressed as
\begin{align}
    \nonumber p_{X_{k+1} \mid N_0, \dots, N_k}&(\V x_{k+1} \mid \V n_0, \dots, \V n_k) = \\
    &\int_{\mathcal X} p_{X' \mid X}(\V x_{k+1} \mid \V x_k)p_{X_k \mid N_0, \dots, N_k}(\V x_k \mid \V n_0, \dots, \V n_k)d\V x_k,
    \label{eq:prediction-equation}
\end{align}
where $\mathcal X$ is the state space of the stimulus. If the state space is countable, then the integral in this equation becomes a sum.

The predictions are a function of the transition density $p_{X' \mid X}$ and the beliefs $p_{X_k \mid N_0 = \V n_0, \dots, N_k = \V n_k}$ at time $k$. Therefore, given the transition and emission densities, we may recursively compute the beliefs at any time $k$ given the sequence of responses $\V n_0, \dots, \V n_k$ by using the update equation to calculate the beliefs as a function of the predictions, and by using the prediction equation to compute the predictions as a function of the previous beliefs. This recursion ultimately completes at the prior over the initial stimulus $p_{X_0}$.

\subsection{Closed-Form Solutions}
\label{sec:closed-form-solutions}

For arbitrary dynamic stimuli and dynamic Poisson populations, the prediction and update equations cannot be evaluated in closed-form. Nevertheless, dynamic Poisson populations do have a particular flexibility when it comes to Bayesian filtering. Before we explain this, first note that the prediction equation (\ref{eq:prediction-equation}) depends on the transition density $p_{X' \mid X}$, but not the emission density $p_{N \mid X}$. This implies that if we know how to solve the prediction equation for a particular dynamic stimulus, then we may apply this solution towards evaluating the Bayes filter for any form of emission density.

Now recall from section \ref{sec:probabilistic-population-codes} that the posterior (\ref{eq:posterior}) computed from the response of a Poisson population (\ref{eq:poisson-likelihood}) and an appropriate prior (\ref{eq:prior}) is a member of an exponential family determined by the form of the tuning curves of the Poisson population (\ref{eq:likelihood}). Moreover, as described in section \ref{sec:stimulus-independent-total-rate}, if the sum of the tuning curves is constant (\ref{eq:tuning-curve-sum}), then the prior is a member of the same exponential family as the posterior. As such, if the sum of the tuning curves of the emission density is constant (\ref{eq:tuning-curve-sum}), and if the predictions at time $k$ (\ref{eq:prediction-equation}) are elements of the same exponential family as the beliefs, then we can compute the exponential family beliefs at time $k$ (\ref{eq:update-equation}) by computing the posterior (\ref{eq:posterior}) as a function of the emission density and the exponential family predictions.

More formally, suppose that $\mathcal M_X$ is the exponential family which matches the tuning curves of the emission density $p_{N \mid X}$, that the sum of the tuning curves of $p_{N \mid X}$ is constant (\ref{eq:tuning-curve-sum}), and that the parameters of the initial prior $p_{X_0} \in \mathcal M_X$ are $\eprms^*$. Then given the initial response $\V n_0$, the natural parameters of the initial belief density $p_{X_0 \mid N_0 = \V n_0} \in \mathcal M_X$ are $\eprms_0 = \iprms_N \cdot \V n_0 + \eprms^*$ in accordance with the posterior of linear probabilistic population codes (\ref{eq:posterior}). Now suppose that the belief density $p_{X_k \mid N_0, \dots, N_k}$ at time $k$ is in $\mathcal M_X$ with natural parameters $\eprms_k$, that the prediction density $p_{X_{k+1} \mid N_0, \dots, N_k}$ at time $k+1$ is also in $\mathcal M_X$, and that there exists a function $\V h$ which computes the natural parameters $\V h(\eprms_k)$ of $p_{X_{k+1} \mid N_0, \dots, N_k}$. Then given the sequence of population responses $\V n_0, \dots, \V n_{k+1}$, we may compute natural parameters of the belief density at time $k+1$ by computing
\begin{equation}
    \eprms_{k+1} = \iprms_N \cdot \V n_{k+1} + \V h(\eprms_k),
    \label{eq:optimal-filter}
\end{equation}
and therefore by induction, we may compute the parameters of the belief density at any $k$.

The question then becomes whether we can find such a function $\V h$. Outside the context of neuroscience, there are two well-known cases for which Bayes filters may be evaluated in closed-form. The first is when the latent variable (i.e. the stimulus) and the observation (i.e. the response) only take on a finite number of values. In this case both the normalization in the update equation and the sum in the prediction equation can be computed brute-force. The second case is when the transition and emission densities are given by linear transformations with additive Gaussian noise. In this case the solution is known as a Kalman filter, and solving the corresponding prediction and update equations may be reduced to straightforward linear algebra \citep{thrun_probabilistic_2005,sarkka_bayesian_2013}.

Now suppose we wish to apply the solutions to the prediction equation provided by these two Bayes filters to computing beliefs given sequences of population responses. The prediction and belief densities are given by categorical densities in the finite-state case, and (multivariate) normal densities in the Kalman filter case, both of which are exponential family densities. This implies that if $\eprms_k$ are the natural parameters of the belief density at any time $k$, then there exists a function $\V h$ such that $\V h(\eprms_k)$ are the natural parameters of the prediction density at time $k+1$, and we may therefore use equation \ref{eq:optimal-filter} to recursively compute the beliefs at all times.

Moreover, even when the prediction densities are not elements of an exponential family, it often remains a good strategy to approximate them with exponential family densities. For example, the most well-known extension of Kalman filtering to nonlinear dynamical systems is the extended Kalman filter \citep{thrun_probabilistic_2005,sarkka_bayesian_2013}. Although the extended Kalman filter does not compute the true predictions, it does compute good exponential family approximations to the true predictions, and we may use this approximation to define the function $\V h$, and thereby approximate a Bayes filter. We apply these techniques later in this paper when we validate our method for learning to compute approximate predictions.

\subsection{Optimal Filtering in Neural Circuits}
\label{sec:optimal-filtering-in-neural-circuits}

In the previous section we demonstrated how to solve the prediction and update equations based on dynamic Poisson populations, and in this section we wish to encode these solutions in a dynamic neural circuit. In section \ref{sec:neural-bayes-rule} we introduced the concept of prior and posterior populations. In the context of Bayesian filtering, we refer to these populations as the prediction population and the filtering population.

\begin{figure}
    \centering
    \includegraphics[scale=0.45]{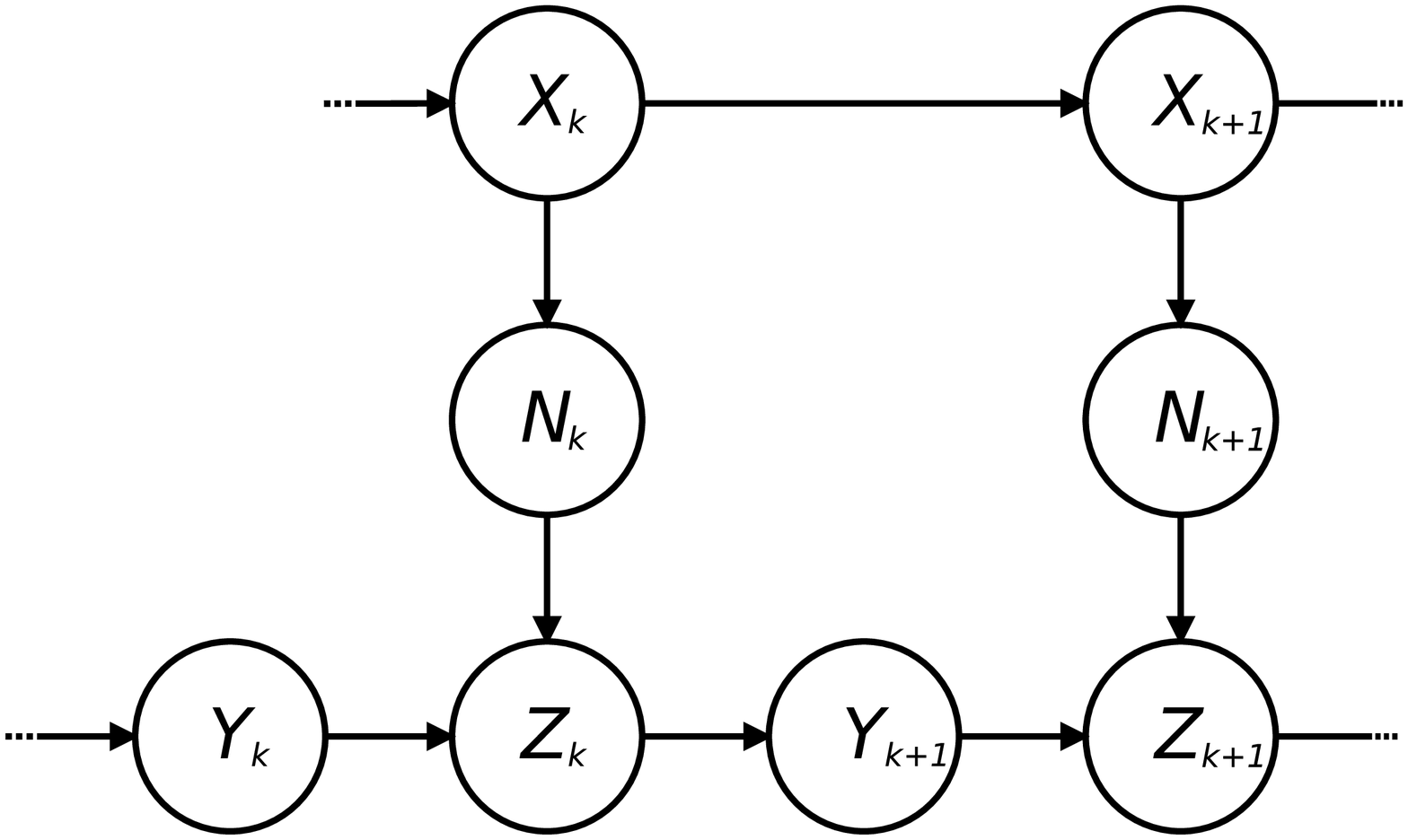}
    \caption{\textbf{Generic Neural Circuit}: Here we depict the graphical representation of the generic neural circuit proposed in this paper. At time $k$, the circuit is composed of a random dynamic stimulus $X_k$, a random dynamic response $N_k$, and the random rates of the prediction and filtering populations, $Y_k$ and $Z_k$. The arrow from $Z_k$ to $Y_{k+1}$ represents the output of the prediction network $\V g$, such that $Y_{k+1} = \V g(Z_k)$. The arrows from $Y_{k+1}$ and $N_{k+1}$ to $Z_{k+1}$ represent the linear combination of $Y_{k+1}$ and $N_{k+1}$, such that $Z_{k+1} = \V A \cdot N_{k+1} + \V B \cdot Y_{k+1}$.}
    \label{fig:coupled-markov-model}
\end{figure}

We begin by defining the components of the dynamic neural circuit. Let $(Y_k)_{k \in \N}$ and $(Z_k)_{k \in \N}$ be the sequence of random firing rates of the prediction and filtering populations. Given the emission density $p_{N \mid X}$ with parameters $\iprms_N$, let $\iprms_Y$, $\iprms_Z$, $\V A$, and $\V B$ be matrices which satisfy equations \ref{eq:population-relation} and \ref{eq:population-relation2}, and let us define the densities encoded by the prediction and filtering populations $Y_k$ and $Z_k$ at any time $k$ as the decoding densities $q_{X \mid Y}$ and $q_{X \mid Z}$ in relations \ref{eq:encoded-prior} and \ref{eq:encoded-posterior}, respectively. Let the rates of the filtering population be $Z_k = \V A \cdot N_k + \V B \cdot Y_k$, let the initial rates of the prediction population $Y_0$ be given by some density $p_{Y_0}$, and finally, for $k > 0$, let the rates of the prediction population be $Y_k = \V g(Z_{k-1})$, where $\V g$ is a neural network which we refer to as the prediction network. We depict the graphical representation of this dynamic neural circuit in figure \ref{fig:coupled-markov-model}.

Let us now consider initial rates of the prediction population $\V y_0$ which satisfy $q_{X \mid Y = \V y_0} = p_{X_0}$, the initial population response $\V n_0$ and the initial rates of the filtering population $\V z_0 = \V A \cdot \V n_0 + \V B \cdot \V y_0$. Since the prior $p_{X_0}$ is equal to $q_{X \mid Y = \V y_0}$ by assumption, the encoded beliefs $q_{X \mid Z = \V z_0}$ are equal to the beliefs $p_{X_0 \mid N_0 = \V n_0}$ in accordance with neural Bayes' rule (\ref{eq:neural-bayes-rule}). Let us now suppose that at an arbitrary time $k$, the rates of the filtering population are $\V z_k$, the subsequent rates of the prediction population are $\V y_{k+1} = \V g(\V z_k)$, and the subsequent response is $\V n_{k+1}$. Let us also suppose that the prediction network $\V g$ has the property that if $q_{X \mid Z = \V z_k}$ is equal to the belief density $p_{X_k \mid N_0 = \V n_0, \dots, N_k \V = \V n_k}$ at time $k$, then $q_{X \mid Y = \V y_{k+1}}$ is equal to the prediction density $p_{X_{k+1} \mid N_0 = \V n_0, \dots, N_k = \V n_k}$ at $k+1$. Since the update equation (\ref{eq:update-equation}) is simply an application of Bayes' rule to the emission and prediction densities, neural Bayes' rule (\ref{eq:neural-bayes-rule}) implies that $q_{X \mid Z = \V z_{k+1}}$ is equal to $p_{X_{k+1} \mid N_0 = \V n_0, \dots, N_{k+1} = \V n_{k+1}}$ where $\V z_{k+1} = \V A \cdot \V n_{k+1} + \V B \cdot \V y_{k+1}$. Therefore, by induction, $\V y_k$ and $\V z_k$ encode the predictions and beliefs at any $k$, and the dynamic neural circuit depicted in figure \ref{fig:coupled-markov-model} exactly implements a Bayes filter for any realization of the system.

The question, again, is whether there in fact exists a neural network $\V g$ which computes encodings of the true predictions. The work of \cite{beck_exact_2007} and \cite{beck_marginalization_2011} answers this question in the affirmative, where the authors derived stochastic differential equations which describe how to encode beliefs in a neural population when the stimulus has either finite-states or is driven by linear dynamics, respectively. Critically, these solutions involve linearly combining the responses of the observation population with encoded predictions at every time step, in accordance with equation \ref{eq:encoded-neural-bayes-rule}.

In the linear dynamical system case, under the assumption that the dynamic population code satisfies equation \ref{eq:tuning-curve-sum}, the prediction network $\V g$ may be expressed as
\begin{equation}
    \V g(\V z) = (\V G^{(2)} \cdot \V z + \V z \cdot \V G^{(3)} \cdot \V z + \V 1 (z^0 - \frac{\V 1 \cdot \V z}{m}))dt + \V z,
    \label{eq:optimal-linear-circuit}
\end{equation}
where $dt$ is the time-step in the time-discretized system. Intuitively, $\V G^{(2)}$ drives the rate of the population in proportion to the linear dynamics, $\V G^{(3)}$ quadratically drives the rate of the population in proportion to the noise in the dynamics, and $z^0$ is a parameter which encourages the component-wise average of the rate process to remain near $z^0$ \citep{beck_marginalization_2011}. Although this is the optimal solution, computing this $\V g$ depends on knowing the parameters of linear the stimulus dynamics, which may not always be available. In the next section we describe how to maximize the likelihood of of a parameterized $\V g$ based the responses of the observation population.

\section{Learning}
\label{sec:learning}

In the previous section we described how to compute beliefs about an unknown, dynamic stimulus with a Bayes filter. Evaluating the Bayes filter requires solving the prediction and update equations, which in turn depend on access to the transition and emission densities, respectively. Although assuming that we can access these densities allows us to define optimal neural circuits directly, this assumption is not ecologically valid. Animals do not, in general, have direct access to emission and transition densities, and must rather learn to implement Bayes filters in their neural circuitry.

In this paper we reduce learning to the optimization of the parameters of a theoretical neural circuit. Different tasks and experimental designs call for training some of these parameters, and leaving others fixed. In our case, we focus on optimizing the parameters of a neural network for computing approximate predictions, and assume that the rest of the parameters of the neural circuit have already been optimized. This scenario applies when dealing with an adult subject with well-developed neural populations for sensation, but no familiarity with the task or dynamic stimulus.

A concrete example of such a task is sequence learning in a psychophysics experiment \citep{clegg_sequence_1998}. Many colour sensitive neurons are found in the visual area V4, which connect further down the ventral stream with neurons in the inferior temporal cortex (ITC) \citep{roe_toward_2012}, which in turn has been found to play a role in sequence learning \citep{meyer_statistical_2011}. We may thus interpret the neural populations in V4 as the observation population, and populations in the ITC as the prediction and filtering populations. In a given sequence learning task, the prediction network is then trained to ensure that the rates of the filtering population in the ITC encode accurate beliefs about the stimulus.

In this section we begin by defining the architecture of our theoretical neural circuit, and introducing the general negative log-likelihood gradient on the parameters of the neural network. We continue by deriving an expression for this gradient, and then introduce contrastive divergence minimization and a novel exponential family approximation in order to descend it.

\subsection{Approximate Filtering in Neural Circuits}

Suppose we know the emission density $p_{N \mid X}$ of a dynamic Poisson population $(X_k, N_k)_{k \in \N}$, and that we wish to train a neural circuit to approximately implement a Bayes filter on responses of the dynamic Poisson population. We take a parametric approach in this paper, which means we must choose a form for the approximate beliefs. We therefore assume that the approximate belief densities are elements of some exponential family $\mathcal M_X$ with sufficient statistic $\V s$ and a constant base measure.

As discussed in sections \ref{sec:probabilistic-population-codes} and \ref{sec:stimulus-independent-total-rate}, in order apply the update equation (\ref{eq:update-equation}) to compute such exponential family beliefs given a response from the observation population, the approximate predictions must be proportional to $e^{\V s(\V x) \cdot \eprms_X + \gamma \sum_{i = 1}^{d_N} f_i(\V x)}$ for some parameters $\eprms_X$. Moreover, as described in section \ref{sec:closed-form-solutions}, since many algorithms for Bayesian filtering yield predictions and beliefs which are in a single exponential family, and since we typically aim to approximate these algorithms as well as possible, we typically assume that the sum of the tuning curves of the observation population is constant such that the approximate prediction densities are also elements of $\mathcal M_X$.

Nevertheless, the brain cannot work with such abstract prediction and belief densities directly. The theoretical neural circuit described in section \ref{sec:optimal-filtering-in-neural-circuits} and depicted in figure \ref{fig:coupled-markov-model} is designed to encode predictions and beliefs with such exponential family forms by way of the prediction and filtering population decoders $q_{X \mid Y}$ (\ref{eq:encoded-prior}) and $q_{X \mid Z}$ (\ref{eq:encoded-posterior}). We therefore assume that the neural circuit considered in this section has the same structure as in section \ref{sec:optimal-filtering-in-neural-circuits}, except now the prediction network $\V g$ is parameterized by $\V \phi$. In our simulations, $\V g$ will be a three layer perceptron with $d_Y$ input neurons, $d_Z$ output neurons, $d_H$ hidden neurons, and $\V \phi$ will be a pair of matrices and biases.

The prediction network $\V g$ computes encodings of the parameters of prediction densities over the stimuli, and so if we could access sample sequences of the stimuli $\V x_0, \dots, \V x_k$, then optimizing $\V g$ would reduce to a regression problem. However, since the purpose of a Bayes filter, approximate or otherwise, is to compute beliefs about unknown stimuli, using these stimuli for training purposes would violate the spirit of the problem. Therefore, we aim instead to optimize $\V g$ based on sequences of population responses $\V n_0, \dots, \V n_k$.

In order to optimize $\V g$ based on sequences of responses, we first define the approximate generative model
\begin{equation}
    q_{XN \mid Y}(\V x, \V n \mid \V y) \propto \frac{e^{\V s(\V x) \cdot \iprms_N \cdot \V n + \V s(\V x) \cdot \iprms_Y \cdot \V y + \V n \cdot \eprms_N}}{n_1! \cdots n_{d_N}!},
    \label{eq:conditional-harmonium}
\end{equation}
which is equal to the harmonium defined in relation \ref{eq:harmonium} where $\iprms_Y \cdot \V y = \eprms_X$. As discussed in sections \ref{sec:probabilistic-population-codes} and \ref{sec:stimulus-independent-total-rate}, this density is the only approximate generative model over stimuli and responses which is consistent with exponential family beliefs, a likelihood given by a population of Poisson neurons, and predictions encoded by $\V y$. Where $q_{N \mid Y}$ is the marginal density of $q_{XN \mid Y}$, and given the sequence of responses $\V n_0, \dots, \V n_k$, we may maximize the likelihood of the parameters $\V \phi$ by following the stochastic negative log-likelihood gradient
\begin{equation}
    -\nabla_{\V \phi}\log q_{N \mid Y}(\V n_k \mid \V y_k),
    \label{eq:cross-entropy-gradient}
\end{equation}
where $\V y_k$ is computed as a function of the neural network and the sequence of responses $\V n_0, \dots, \V n_{k-1}$ \citep[see][]{welling_exponential_2004,bengio_learning_2009}.

\subsection{Computing the Gradient}
\label{sec:computing-the-gradient}

As shown in \cite{welling_exponential_2004}, the component partial derivatives of the negative log-likelihood gradient of $\eprms_X$ for the harmonium in relation \ref{eq:harmonium} are
\begin{equation}
    -\partial_{\eprms_X} \log p_N(\V n)= \E_p[\V s(X) \mid N = \V n] - \E_p[\V s(X)],
    \label{eq:harmonium-gradient}
\end{equation}
where the random variables are distributed according to $p_{XN}$. Given a sequence of responses $\V n_0, \dots, \V n_k$ and the corresponding rates of the prediction population $\V y_0, \dots, \V y_k$, if we consider the derivative of the negative log-likelihood with respect to the component parameters $\phi_i$ of $\V \phi$, and define the biases as $\eprms_X = \iprms_Y \cdot \V y_k$, we may apply the chain rule to combine derivative \ref{eq:harmonium-gradient} with the partial derivative $\partial_{\phi_i} \V y_k$. By taking the expectations with respect to $q_{XN \mid Y}$ and adding the dependencies on $Y_k$, we may then write the partial derivatives as
\begin{align}
    \nonumber -\partial_{\phi_i} &\log q_{N \mid Y}(\V n_k \mid \V y_k) = \\
    &(\E_q[\V s(X) \mid N = \V n_k, Y = \V y_k ] - \E_q[\V s(X) \mid Y_k = \V y_k]) \cdot \iprms_Y \cdot \partial_{\phi_i} \V y_k.
    \label{eq:prediction-gradient1}
\end{align}
Observe that this gradient is zero when the predictions of the network match the posterior, and is thus a form of prediction error.

Because $\V y_k$ depends recurrently on $\V y_0, \dots, \V y_{k-1}$, we must recursively apply the chain rule to the gradient $\partial_{\phi_i} \V y_k$ until $\V y_0$ is reached. This leads to the algorithm known as backpropagation-through-time \citep{werbos_backpropagation_1990,sutskever_recurrent_2009,makin_recurrent_2016} for computing the recursive gradient of $\V y_k$. Unfortunately, backpropagation-through-time is known to be problematic \citep{bengio_learning_1994,pascanu_difficulty_2013}, and introduces complexities into the gradient calculation that we wish to avoid in this paper.

In section \ref{sec:optimal-filtering-in-neural-circuits} we showed how to define optimal and near-optimal filtering populations such that $(Z_k)_{k \in \N}$ loses (nearly) no information about the stimulus over time. This implies that $(Z_k)_{k \in \N}$ is (approximately) Markov, and so at the optimal parameters $\V \phi$ of $\V g$, $Y_k$ is (nearly) independent of $Z_{k-j}$ for $j > 1$. This suggests that we may ignore the long-range dependencies in the gradient and still hope to find good parameters. We therefore consider a one-step approximation to the true derivative of $\V y_k$, and assume that the rates of the filtering population at the previous time $\V z_{k-1}$ are independent of the parameters $\V \phi$. This allows us to express the components of gradient \ref{eq:cross-entropy-gradient} as
\begin{align}
    \nonumber -\partial_{\phi_i} &\log q_{N \mid Y}(\V n_k \mid \V y_k) = \\
    &(\E_q[\V s(X) \mid N = \V n_k, Y = \V y_k] - \E_q[\V s(X) \mid Y_k = \V y_k]) \cdot \iprms_Y \cdot \partial_{\phi_i} \V g(\V z_{k-1}),
    \label{eq:prediction-gradient2}
\end{align}
and thereby reduce the problem of maximizing the likelihood of the parameters $\V \phi$ to computing equation \ref{eq:harmonium-gradient} and the partial derivatives $\partial_{\phi_i} \V g$ at $\V z_{k-1}$. When $\V g$ is a multilayer perceptron, we may apply standard backpropagation to compute these derivatives \citep{rumelhart_learning_1986}. The last remaining problem is therefore to compute the expectations $\E_q[\V s(X) \mid N = \V n_k, Y = \V y_k]$ and $\E_q[\V s(X) \mid Y = \V y_k]$.

\subsection{Approximating the Harmonium Expectations}
\label{sec:approximating-the-harmonium-expectations}

Computing the first conditional expectation $\E_q[\V s(X) \mid N = \V n_k, Y = \V y_k]$ is not challenging, as $q_{X \mid N = \V n_k, Y = \V y_k}$ is given by the decoder of the filtering population $q_{X \mid Z = \V A \cdot \V n_k + \V B \cdot \V y_k}$ (\ref{eq:encoded-posterior}) in accordance with neural Bayes' rule (\ref{eq:neural-bayes-rule}). By design $q_{X \mid Z}$ is always an element of some exponential family, and for many exponential families the expected value of the sufficient statistics can be computed exactly, or in the very least can be approximated by sampling. On the other hand, $\E_q[\V s(X) \mid Y = \V y_k]$ is determined by the encoded prediction density $q_{X \mid Y}$ (\ref{eq:encoded-prior}), which cannot, in general, be trivially evaluated or sampled. In this paper we consider two strategies for approximating this expectation.

Firstly, the expectations of the marginal densities of an exponential family harmonium may be approximated by Gibbs sampling \citep{geman_stochastic_1984, roberts_geometric_1994}. In this context, gibbs sampling involves constructing a Markov chain through recursive sampling of the densities $q_{N \mid X, Y}$ and $q_{X \mid N, Y}$ based on an arbitrary initial response $\V n^0$. It can be shown that the sample stimuli and responses generated after many such iterations are distributed approximately according to the density $q_{XN \mid Y}$. Since it often takes a long time for this Markov chain to converge, the contrastive divergence gradient was developed as an alternative to the negative log-likelihood gradient \citep{hinton_training_2002,bengio_justifying_2009}. Approximating the contrastive divergence gradient leads to a similar approximation scheme as Gibbs sampling, however, instead of an arbitrary starting condition which we wish the sampler to forget, we let $\V n^0 = \V n_k$, which allows a useful gradient to be calculated after a handful of iterations.

Secondly, if the sum of the tuning curves of the emission density $p_{N \mid X}$ are constant (\ref{eq:tuning-curve-sum}), then the decoder of the prediction population $q_{X \mid Y}$ is in the same exponential family as the decoder of the filtering population $q_{X \mid Z}$, and therefore $\E_q[\V s(X) \mid Y = \V y_k]$ can be evaluated as easily as $\E_q[\V s(X) \mid N = \V n_k, Y = \V y_k]$. In particular, where $\V \tau$ is the coordinate transform from the natural parameters to the expectation parameters of the exponential family of $\V s$ \citep{amari_methods_2007,wainwright_graphical_2008,nielsen_statistical_2009}, equation \ref{eq:harmonium-gradient} is then given by
\begin{equation}
    \E_p[\V s(X) \mid N = \V n] - \E_p[\V s(X)] = \V \tau (\iprms_N \cdot \V n_k + \eprms_X) - \V \tau(\eprms_X).
    \label{eq:closed-form-expectations}
\end{equation}
In the case of the 1-dimensional Gaussian tuning curve, $\V \tau$ may be computed in closed-form, and is given by
\begin{equation}
    \V \tau(\theta_{X,1},\theta_{X,2}) = \left(-\frac{\theta_{X,1}}{2\theta_{X,2}},\frac{\theta_{X,1}}{4\theta_{X,2}} - \frac{1}{2\theta_{X,1}}\right).
\end{equation}
Nevertheless, equation \ref{eq:tuning-curve-sum} is often only approximately satisfied, and so this strategy may not always be optimal. In the subsequent section, we compare the performance of this exponential family approximation with contrastive divergence minimization.

\section{Simulations}
\label{sec:simulations}

In this section we describe in detail how to apply the methods we have developed in three simulated experiments. In each experiment we aim to understand how a subject learns to compute accurate beliefs about an unknown dynamic stimulus with a neural circuit composed of an observation population which generates responses to stimuli, a prediction population which encodes approximate predictions, a filtering population which encodes approximate beliefs, and a prediction network which computes rates of the prediction population as a function of the rates of the filtering population. In the first half of this section we describe the details of the neural circuits, the training procedures, and the validation procedures, which are more or less the same across all experiments. In the second half of this section we present the results of the three simulated experiments.

\subsection{Methods}

The theoretical neural circuits we consider are composed of the observation, prediction, and filtering populations, and the prediction network. The parameters of the observation population are the parameters of the emission density $\iprms_N$ and $\eprms_N$, and the observation population recoder $\V A$; the parameters of the prediction population are the decoding matrix $\iprms_Y$, the prediction population recoder $\V B$, and the initial rates $\V y_0$; and the parameters of the filtering population are the decoding matrix $\iprms_Z$. We set these parameters of the three neural populations by hand. The remaining parameters are the parameters $\V \phi$ of the prediction network $\V g$, which we optimize by stochastic gradient descent on the negative log-likelihood of $\V \phi$ given sequences of population responses.

In order to test what kind of population codes are likely used by the brain, we propose two candidate neural circuits which differ in how the prediction and filtering populations encode probabilities, as defined by $\iprms_Y$ and $\iprms_Z$. Moreover, when training the prediction network in each circuit, we apply and compare contrastive divergence minimization \citep{hinton_training_2002} and the exponential family approximation to the negative log-likelihood gradient (\ref{eq:closed-form-expectations}), leading to a total of four sub-experiments in each experiment. Because we keep the stimuli mathematically simple, we may then validate the learned filter against the corresponding optimal, or mostly optimal, filter.

All simulations presented in this paper were developed in Haskell, and are available at the repository of the author at \emph{hub.darcs.net/alex404/goal}.

\subsubsection{Population Parameters}

In each simulation we define the parameters of the emission density $\iprms_N$ and $\eprms_N$ by first defining the gain $\gamma$ and tuning curves $(f_i)_{i=1}^{d_N}$ such that the sum of the tuning curves is (approximately) constant.  We then equate the likelihood of the Poisson population (\ref{eq:poisson-likelihood}) with its exponential family form (\ref{eq:likelihood}) and solve for $\iprms_N$ and $\eprms_N$. We define the decoding matrix $\iprms_Y$ as equal to the decoding matrix $\iprms_Z$. This implies that the prediction population recoder $\V B = \V I$, since the population codes of the prediction and filtering populations are the same. In this case we may intuitively think of the prediction and filtering populations as a single population for encoding approximate beliefs. Finally, since the sum of the tuning curves of the emission density is constant (\ref{eq:tuning-curve-sum}), we set the initial rates of the prediction population to $\V y_0 = \V 0$, such that the prediction population initially encodes a flat prior over the stimulus.

We define the parameters of the filtering population $\iprms_Z$ in one of two ways. The first is by setting $\iprms_Z = \iprms_N$, which we refer to as the \emph{naive} code. In the case, the observation population recoder is given by $\V A = \V I$. We refer to the second code as the \emph{orthogonal} code, based on the code presented in the supplementary material of \cite{beck_marginalization_2011}. In this case we construct $\iprms_Z$ from a set of mutually orthogonal rows, which are also orthogonal to the vector of ones, such that $\iprms_{Z,(i)} \cdot \iprms_{Z,(j)} = 0$ for $i \neq j$, and $\iprms_{Z,(i)} \cdot \V 1 = 0$. In this case $\iprms_Z \cdot \V 1 = \V 0$, which implies that for any rates of the filtering population $\V z$, the rates $\V z + c \V 1$ encode the same beliefs for any scalar $c$. The details of constructing $\iprms_Z$, as well as a corresponding observation population recoder $\V A$ which satisfies equation \ref{eq:population-relation}, can be found the supplementary material of \cite{beck_marginalization_2011}.

For the naive code, because $\iprms_N = \iprms_Y = \iprms_Z$, all three neural populations in the circuit have the same number of neurons. In the case of the orthogonal code, because $\iprms_Y = \iprms_Z$, the prediction and filtering populations continue to have the same number of neurons. Although we could set the number of neurons in these populations to be different from the number in the observation population, in order to ensure that differences in circuit performance are not simply due to differences in the number of parameters, we continue to define $\iprms_Z$ to have the same number of columns as $\iprms_N$. This implies that in all the cases we consider, $d_N = d_Y = d_Z$, which is to say that the observation, prediction, and filtering populations always have the same number of neurons. For the number of neurons in each experiment, and a summary of all simulation parameters, see table \ref{tab:parameters}.

\begin{table}
    \begin{center}
        \begin{tabular}{| c | c | c | c |}
            \hline
            Parameter & Experiment 1 & Experiment 2 & Experiment 3 \\ \hline
            $d_N = d_Y = d_Z$ & 10 & 10 & 20 \\ \hline
            $d_H$ & 100 & 200 & 500 \\ \hline
            $n_{t}$ & $10,000$ & $10,000$ & $20,000$ \\ \hline
        \end{tabular}
    \end{center}
    \caption{\textbf{Summary of Circuit and Training Parameters}: In this table we show the simulation parameters which change in each experiment. These are the sizes of the observation, prediction, and filtering populations $d_N$, $d_Y$, $d_Z$, respectively; the number of hidden neurons in the prediction network $d_H$; and the number of steps in the training simulation $n_t$. In all experiments, $\iprms_Z = \iprms_Y$ and $\V y_0 = \V 0$; in the naive circuit $\iprms_Z = \iprms_N$, and in the orthogonal circuit $\iprms_Z$ is constructed from orthogonal rows which are also orthogonal to $\V 1$. Where $i_e$ is the epoch, the parameters of the Adam algorithm are $\alpha = 0.00005 \cdot 1.25^{-(i_e-1)}$, $\beta_1 = 0.9$, $\beta_2 = 0.999$, and $\epsilon = 10^{-8}$, and contrastive divergence is run with $i_e$ contrastive divergence steps.}
    \label{tab:parameters}
\end{table}

\subsubsection{Prediction Network and Training Procedure}

In all experiments we define the prediction network $\V g$ as a 3-layer perceptron with $d_H$ hidden neurons, where the number of input and output units is determined by the number of neurons in the filtering population. We define the activation functions of the neural network to be a sigmoid activation function in the hidden layer, and an exponential activation function in the output layer. We use the exponential function on the output layer in order to ensure that the rates computed by the prediction network are always positive. This is especially important in the case of the naive code, as negative rates cannot be reliably decoded by $\iprms_N$.

We denote the parameters of $\V g$ by $\V \phi$, such that $\V \phi$ is composed of a pair of matrices and biases. We thus compute the gradient in equation \ref{eq:prediction-gradient2} by applying backpropagation to compute $\partial_{\phi_i} \V g(\V z_{k-1})$ \citep{rumelhart_learning_1986}, and one of the two methods proposed in section \ref{sec:approximating-the-harmonium-expectations} to compute the expectations in equation \ref{eq:harmonium-gradient}. Given the proposed neural circuits and gradient approximation schemes, we run four parallel simulations in every experiment, by applying either contrastive divergence minimization (CD) or the exponential family approximation (EF) to computing the stochastic gradient (\ref{eq:cross-entropy-gradient}), in order to train the neural circuits based on either the naive (NV) or orthogonal (OT) population codes. We denote these four simulations and corresponding circuits by NV-EF, NV-CD, OT-EF, and OT-CD.

In each experiment we train each neural circuit over the course of twenty epochs, where each epoch is composed of a training simulation of $n_t$ steps. During the training simulation, where $i_e$ is the number of the current epoch, we reset the rates of the prediction population $Y_k$ to $\V 0$ every $(i_e-1)^2$ number of steps. We do this because newly initialized prediction networks $\V g$ tend to be unstable, in that recursively evaluating $Z_k$ for large $k$ tends to result in rates which diverge and fail to encode accurate beliefs. By first training $\V g$ on shorter, stable paths, we may avoid this problem and better maximize the likelihood of the parameters $\V \phi$.

When updating $\V \phi$, we apply the Adam algorithm in order to dynamically adapt the step size of the gradient descent \citep{kingma_adam:_2014}. In every epoch we define the initial learning rate to be $\alpha = \frac{0.00005}{1.25^{i_e-1}}$, we set the momentum parameters of the Adam algorithm to $\beta_1 = 0.9$ and $\beta_2 = 0.999$, and the regularizer to $\epsilon = 10^{-8}$. Finally, when applying contrastive divergence minimization, we also set the number of contrastive divergence steps equal to the epoch number $i_e$.

\subsubsection{Validation}

After each training epoch we validate the trained neural circuits on a simulation of $n_v = 200,000$ steps. We compute the sequence of rates $\V z_0, \dots, \V z_{n_v}$ as a function of the sequence of validation responses $\V n_0, \dots, \V n_{n_v}$ without resetting the rates of the prediction population. Where $\varepsilon$ is a function which measures error given a stimulus and the natural parameters of a belief density, we then compute the average of the error measure $E_Z = \sum_{i=0}^{n_v} \frac{\varepsilon(\V x_i, \iprms_Z \cdot \V z_i)}{n_v}$.

By computing the average error $E_{Opt} = \sum_{i=0}^{n_v} \frac{\varepsilon(\V x_i, \eprms_k)}{n_v}$ on the belief parameters $\eprms_k$ of the closed-form filters described with equation \ref{eq:optimal-filter}, we compute a lower-bound on the error of the trained circuits. Conversely, since any useful filter must provide more information about the stimulus then the instantaneous responses, we compute $E_N = \sum_{i=0}^{n_v} \frac{\varepsilon(\V x_i, \iprms_N \cdot \V n_i)}{n_v}$ to provide a performance upper-bound. Finally, by computing the ratio $r = \frac{E_Z - E_N}{E_{Opt} - E_N}$, we may express the performance of the neural circuit in question as a percentage of the distance achieved from the upper- to the lower-bound.

In the first two experiments we validate our model by computing the average negative log-likelihoods of the approximate beliefs given the true stimuli; that is, $\varepsilon(\V x, \eprms) = -\log q(\V x)$, where $q$ is the exponential family density described in relation \ref{eq:exponential-family} with parameters $\eprms$. In these experiments we can compute the true beliefs of the Bayes filter based on knowledge of the stimulus dynamics. The true beliefs have minimal negative log-likelihood, and the smaller the difference between the average negative log-likelihood of the approximate beliefs and true beliefs, the better our neural circuit has implemented a Bayes' filter.

In the third experiment we model a two-dimensional stimulus with a stochastic pendulum, where the first dimension in the angle and the second is the angular velocity. We define the tuning curves of the observation population as the concatenation of a set of von Mises and normal tuning curves, and the exponential family density which corresponds to these tuning curves is the product density of a von Mises density and a normal density. Because von Mises densities cannot be computed in closed-form, we cannot evaluate the corresponding negative log-likelihoods. Therefore, we instead take our error measure to be the average squared error of the mean of the belief density from the true stimulus, such that $\varepsilon(q,\dot q,\eprms_{vM},\eprms_n) = \frac{(q - \mu_1(\eprms_{vM}))^2 + (\dot q - \mu_2(\eprms_n))^2}{2}$, where $(\mu_{vM}(\eprms_{vM}),\mu_n(\eprms_n))$ is the mean of of the bivariate von Mises-normal density with parameters $\eprms = (\eprms_{vM}, \eprms_n)$.

Since pendula are nonlinear dynamical systems and the tuning curves are von Mises, we cannot exactly compute the true Bayes filter for the third experiment. An extended Kalman filter (EKF) is an algorithm for approximate filtering of nonlinear dynamical systems, however the predictions of an EKF can only be computed as a function multivariate normal beliefs, not von Mises-normal beliefs. Nevertheless, when the concentration parameter of the von Mises density is high, it is approximately equal to the inverse variance of the density, and the von Mises density is approximately normal. We therefore compute approximate EKF predictions in the following manner.

First note that if the von Mises-normal beliefs at some time are given by some natural parameters $\eprms$, then we may equivalently express this density with the parameters $(\mu_{vM}, \kappa, \mu_n, \sigma^2)$, where $\kappa$ is the concentration of the von Mises density, and $\sigma^2$ is the variance of the normal density. We convert this density into a multivariate normal density by setting the mean $\mu_{EKF}$ of the multivariate normal to $\mu_{EKF} = (\mu_{vM}, \mu_n)$, and the covariance matrix $\Sigma_{EKF}$ to a diagonal matrix with first component $\Sigma_{EKF,(1,1)} = 1/\kappa$ and second component $\Sigma_{EKF,(2,2)} = \sigma^2$. After computing the multivariate normal predictions with parameters $\mu^*_{EKF}$ and $\Sigma^*_{EKF}$ as a function of the parameters $\mu_{EKF}$ and $\Sigma_{EKF}$ with the EKF, we convert the multivariate normal predictions back into von Mises-normal predictions by setting $(\mu^*_{vM},\mu^*_n) = \mu^*_{EKF}$, $\kappa^* = \Sigma_{EKF,(1,1)}^*$, and $(\sigma^2)^* = \Sigma^*_{EKF,(2,2)}$. After computing the approximate predictions, we apply Bayes' rule by evaluating equation \ref{eq:optimal-filter} in the usual manner. As we later show, when the concentration of the von Mises density is high, this provides an effective filter for this nonlinear, von Mises-normal system.

Finally, in the last two experiments we also estimate the tuning curves of the hidden layer of the prediction network with respect to the stimuli. We estimate these tuning curves by simulating the trained neural circuits for $n_v$ steps, and then sorting these steps into bins, where each bin contains the activity of the hidden layer of the prediction network when the stimulus is near a particular stimulus value. We then average the rate of each hidden neuron in each bin, in order to estimate the mean activity of the hidden neuron given the stimulus which corresponds to the bin.

\subsection{Results}

In this section we simulate three experiments which model how a neural circuit learns to implement a Bayes filter. The three stimuli are colour sequences which we model as a finite-state Markov chain; the position of a mouse on a track which we model as a 1-dimensional linear dynamical system; and the angle and angular velocity of a human arm, which we model as a pendulum. In the first experiment the tuning curves of the observation population are designed to ensure that the sum of the tuning curves is exactly constant. In the second experiment the tuning curves are Gaussian, and in the third experiment the tuning curves are a product of a von Mises and a Gaussian tuning curve. In both the second and third experiments the tuning curves tile the space of the stimulus so that the sum of the tuning curves is approximately constant.

\subsubsection{Colour Sequence Learning}
\label{sec:colour-sequence-learning}

In this simulated experiment we imagine that subjects are shown sequences of colours drawn from red, green, and blue. The colours are described by a Markov chain, such that each colour has a certain probability of appearing based on the previously seen colour. Subjects must learn to predict the sequence as well as possible. We assume that the stimuli change quickly, so that subjects do not always perceive the stimulus before it transitions to the next stimulus value. We consider how this problem might be solved in the ventral stream, and model colour-sensitive neurons in the visual area V4 with the observation population, and sequence-learning neurons in the inferior temporal cortex with the prediction and filtering populations \citep{roe_toward_2012}.

For simplicity, let us denote the three colour values by $r$, $g$, and $b$. The transition probabilities of the Markov chain are
\begin{align*}
    &p_{X' \mid X}(r \mid r) = p_{X' \mid X}(b \mid b) = 0.8, && p_{X' \mid X}(g \mid g) = 0.5, \\
    &p_{X' \mid X}(r \mid g) = p_{X' \mid X}(b \mid g) = 0.25, && p_{X' \mid X}(g \mid r) = p_{X' \mid X}(g \mid b) = 0.15, \\
    &p_{X' \mid X}(b \mid r) = p_{X' \mid X}(r \mid b) = 0.05.
\end{align*}
Intuitively, blue tends to stay blue and red tends to stay red, whereas green is a relatively transitory state. Moreover, red and blue tend to first transition through green before reaching blue and red, respectively.

We assume that the observation population has $d_N = 10$ neurons, and that the gain $\gamma = 1$. We define the tuning curve of neuron $i$ given blue as $f_i(b) = e^{0.4(i-1) - 5}$, given red as $f_i(r) = f_{10-i}(b)$ and given green as $f_i(g) = \frac{1}{n}\sum_{i=1}^{10} f_i(b)$. This construction ensures that equation \ref{eq:tuning-curve-sum} is satisfied exactly. Intuitively, the low-index neurons of the observation population respond to red, the high-index neurons respond to blue, and the observation population responds with a uniform pattern of activity to green, which provides little information about the true colour. Finally, we set the number of hidden neurons in the prediction network to $d_H = 100$.

\begin{figure}[t!]
    \begin{minipage}{0.5\textwidth}
        \includegraphics[width=\textwidth]{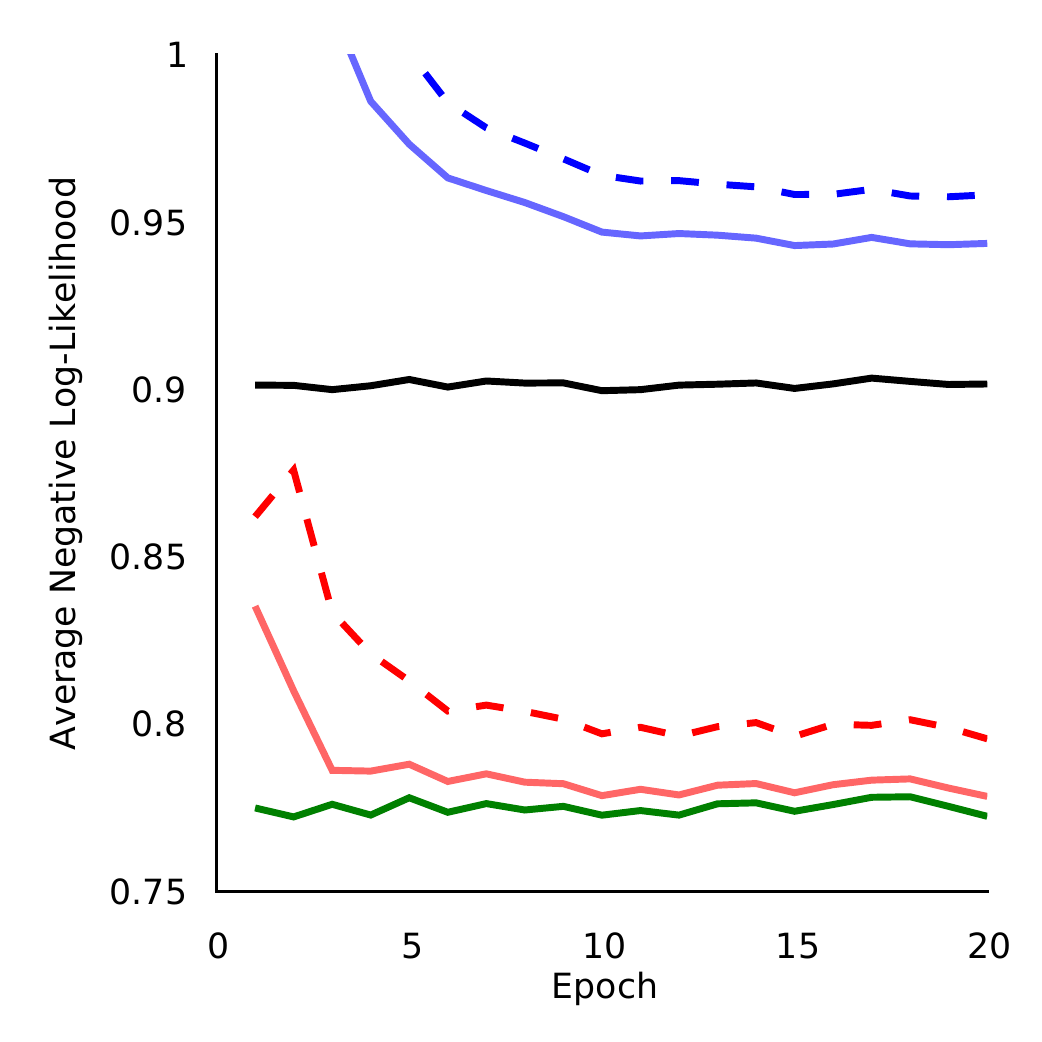}
    \end{minipage}
    \begin{minipage}{0.5\textwidth}
        \includegraphics[width=\textwidth]{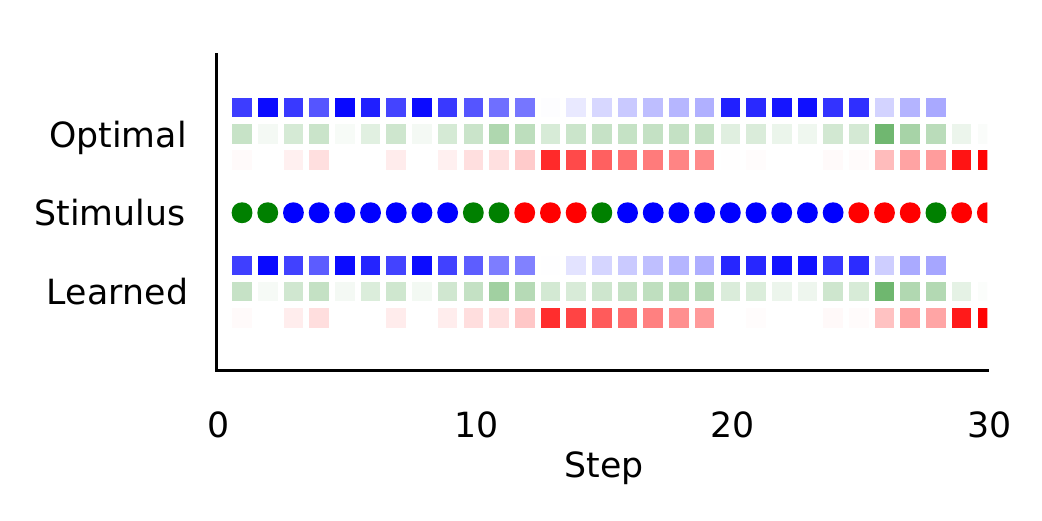}
        \raggedleft \includegraphics[width=0.92\textwidth]{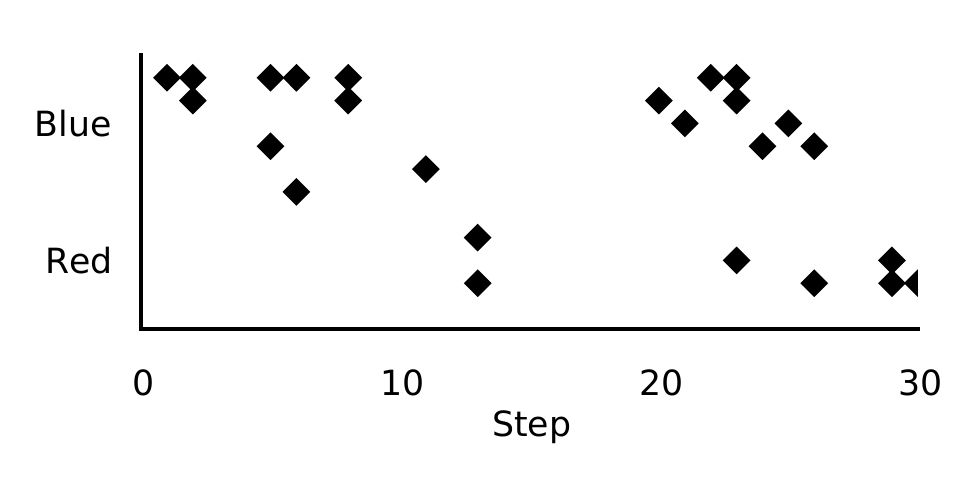}
    \end{minipage}
    \caption{\textbf{Colour Sequence Training and Simulation}: Here we depict the training of the proposed neural circuits, and a simulation with the OT-EF circuit. \emph{Left}: The average negative log-likelihood of the approximate beliefs given the stimuli over each epoch. We display the descent of the NV-EF circuit (blue), the NV-CD circuit (dashed blue), the OT-EF circuit (red) and the OT-CD circuit (dashed red). We also depict the baseline (black) provided by the population responses and the optimum (green) computed by the discrete Bayes filter. \emph{Top Right}: A simulation from the Markov chain (coloured circles), as well as the learned an optimal filters. The beliefs of the optimal filter (top) and the learned filter (bottom) are indicated by the opacity of a colour, which corresponds to the inferred probability of the stimulus value. \emph{Bottom Right}: The responses of the observation population over the 30 steps of the simulation. Spikes (black diamonds) from a particular neuron are arranged along the x-axis in accordance with the neuron index.}
    \label{fig:neural-hmm}
\end{figure}

We depict the results of the simulations of the NV-EF, NV-CD, OT-EF, and OT-CD circuits in figure \ref{fig:neural-hmm}. As displayed in the left panel, the circuit which best approximates the true beliefs of the Bayes filter is the orthogonal circuit trained with the exponential family approximation of the stochastic cross-entropy gradient (solid red), which achieves $r = 95.4\%$ of the performance of the Bayes filter. In this experiment, because equation \ref{eq:tuning-curve-sum} is exactly satisfied, it is unsurprising that the EF gradient produces the best results, as it is in fact equal to the true stochastic gradient. It is surprising, however, that the choice of population code has such a dramatic effect on the learning. Where the orthogonal circuits more or less completely recover the true beliefs, the naive circuits cannot even surpass the baseline provided by the responses.

In the right panels of figure \ref{fig:neural-hmm} we display 30 steps of a simulation from this system. In the top right panel we use the opacities of coloured squares to show the dynamic beliefs of both the optimal and OT-EF filter, and one can see how the beliefs of the two filters are nearly identical. In the bottom right panel we show the corresponding population responses. In total, the observation population spikes 25 times over the course of the simulation. Both filters initially recognize that the stimulus is blue. However, in the middle of the simulation when the stimulus changes to red and back to blue again, the filters cannot recognize this, as no spikes reveal this transition.

\subsubsection{Self-Localization}
\label{sec:self-localization}

In this simulated experiment we imagine that a mouse is confined to a one-dimensional track, and explores the local track while avoiding straying too far from its home position. We model the dynamics of the position of the mouse with a stochastic, one-dimensional, linear dynamical system. We wish to understand how the mouse learns to track its position in a novel environment with place cells in the hippocampus, given noisy position estimates provided by visual cues \citep{mcnaughton_path_2006}. We model place cells with the filtering population, and cue-sensitive cells with the observation population.

Since the position of the mouse is a continuous-time variable, we describe it with the linear stochastic differential equation
\begin{equation*}
    dX_t = a X_tdt + b dW_t,
\end{equation*}
where $W_t$ is a Wiener process. Where $h$ is the time-step, this implies that the transition density $p_{X' \mid X}$ of the time-discretized system at $x$ is a normal density with mean $x + hax$ and variance $hb^2$. In our case we let $a = -1$, $b=1$, and $h=0.02$. We then define the observation population to have $d_N = 10$ neurons with the 1-d Gaussian tuning curves defined in equation \ref{eq:gaussian-tuning-curve}, with gain $\gamma = h\cdot 100 = 2$, preferred stimuli $x^0_i$ distributed evenly over the interval $[-7,7]$, and variance $\sigma^2 = 2$. Finally, we set the number of hidden neurons of the prediction network to $d_H = 200$.

\begin{figure}[t!]
    \begin{minipage}{0.5\textwidth}
        \includegraphics[width=\textwidth]{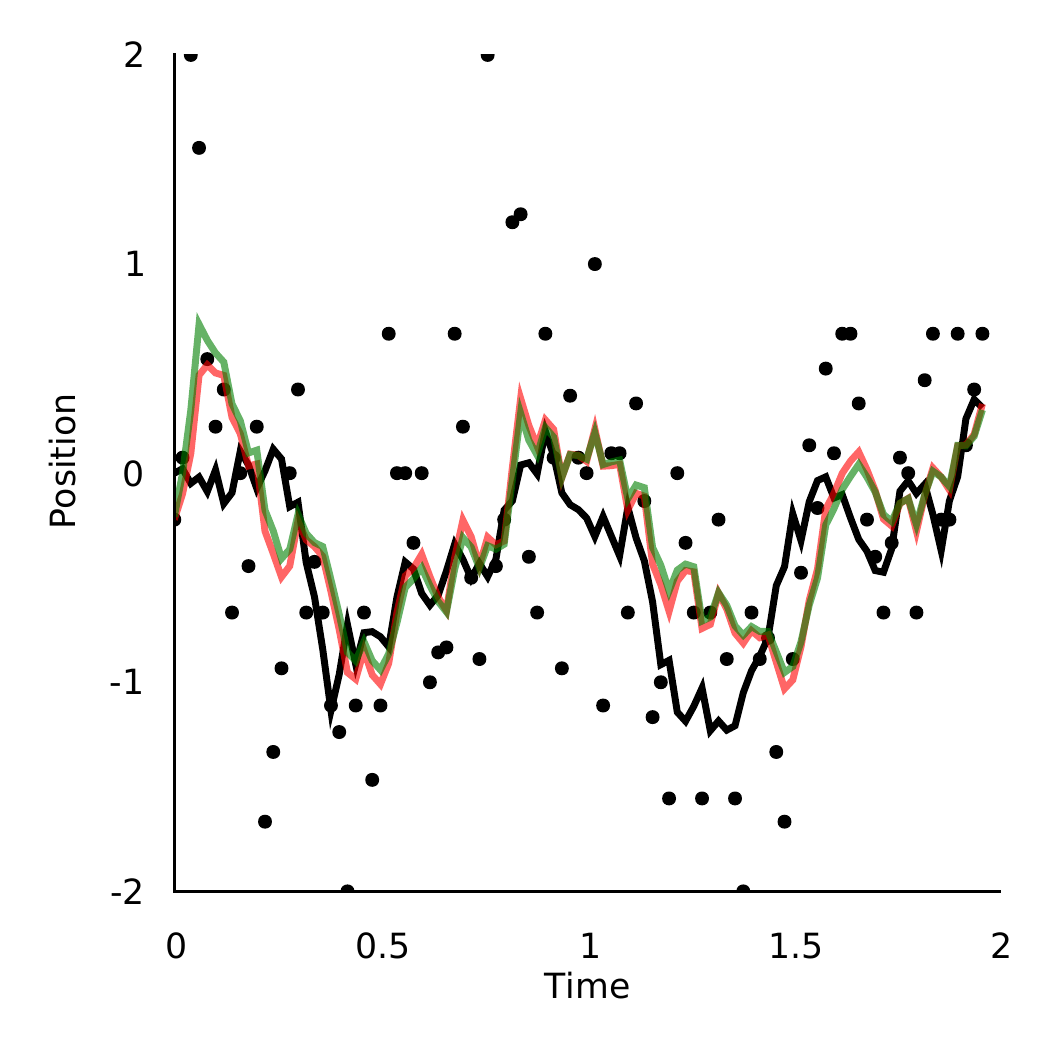}
    \end{minipage}
    \begin{minipage}{0.5\textwidth}
        \includegraphics[width=\textwidth]{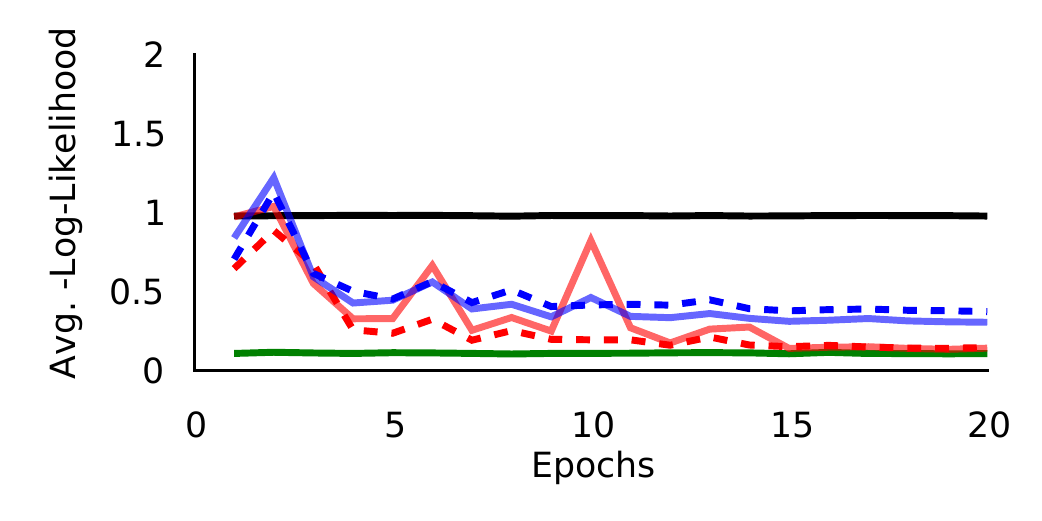}
        \includegraphics[width=\textwidth]{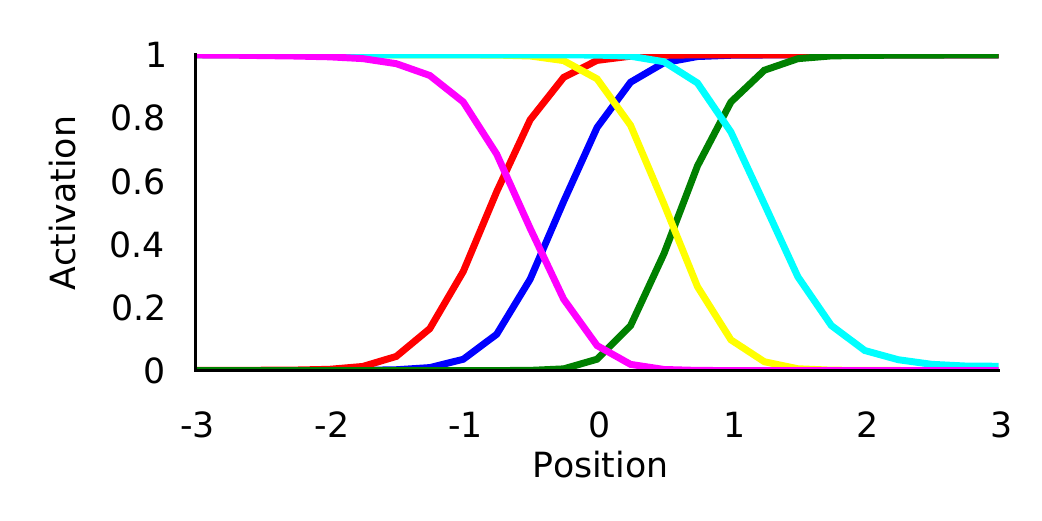}
    \end{minipage}
    \caption{\textbf{Self-Localization Training and Simulation}: Here we depict the training of the proposed neural circuits, and simulations with the OT-EF circuit. \emph{Top Right}: The descent of the negative log-likelihood using the same colour scheme as in figure \ref{fig:neural-hmm}. \emph{Left}: A simulation from the dynamical system, where we depict the stimulus (black line), and the dynamic mean of the response posteriors (black dots), the optimal belief density (green line), and the learned belief density (red line). \emph{Bottom Right}: Six tuning curves from the hidden layer of the prediction network.}
    \label{fig:attractor}
\end{figure}

We depict the results of the four simulations in figure \ref{fig:attractor}. As depicted in the top right panel, the orthogonal circuit trained with the exponential family approximation (red) best approximates the optimal filter, achieving $r = 96.0\%$ of the performance of the optimal filter, which is slightly better than the OT-CD circuit. In the left panel we display a 2 second simulation from the system. The black dots indicate the mean of the posterior $p_{X \mid N = \V n_k}$ of each response $\V n_k$ from equation \ref{eq:posterior} under the assumption of a flat prior. The mean of the optimal beliefs (green line) given these responses is very close to the true stimulus, and the mean of the learned beliefs (red line) is nearly identical to the optimum. In the bottom right panel we display six approximate tuning curves from the hidden layer of the trained multilayer perceptron $\V g$, which we find to have learned sigmoid tuning curves over the stimuli.

\subsubsection{Proprioception}
\label{sec:proprioception}

In this final simulated experiment we imagine that a human is trying to optimize its forward model of the swing of its arm. We focus on the role of the cerebellum in proprioception, and assume that Purkinje cells in the cerebellum receive information about the angle and angular velocity of the shoulder from proprioceptors, and use this information to drive a forward model of arm position \citep{kawato_internal_2003,franklin_computational_2011}. We model the neural populations in the cerebellum with the prediction and filtering populations, and the proprioceptors with the observation population.

For simplicity, we assume that the arm may be described by a single rigid body at a joint, and that the subjects use random motions of the arm in order to explore its dynamics. We therefore model the arm as a stochastic pendulum, which is a two-dimensional stochastic process over the angular position $q$, and the angular velocity $\dot q$. We define the discrete-time transition dynamics $p_{X' \mid X}$ of the stochastic pendulum at $\V x = (q,\dot q)$ as a multivariate normal density with mean $\V x + h \V a(\V x)$ and covariance matrix $h^2\V \Sigma$, where $h$ is the time-step. The function $\V a$ is known as the drift, and is given by
\begin{align*}
    a_1(q,\dot{q}) &= \dot{q}, \\
    a_2(q,\dot{q}) &= -g\sin(q)-c\dot{q},
\end{align*}
where $g = 9.81$ is the gravitational constant and $c = 0.1$ is the coefficient of friction. We define the covariance matrix of the process by
\begin{align*}
    \V \Sigma_{(1,1)}(q,\dot{q}) &= \V \Sigma_{(1,2)}(q,\dot{q}) = \V \Sigma_{(2,1)}(q,\dot{q}) = 0, \\
    \V \Sigma_{(2,2)}(q,\dot{q}) &= \sigma^2_{\dot{q}},
\end{align*}
where $\sigma^2_{\dot{q}} = 1$ is the variance of the noise process. By restricting the noise to the velocity, we may interpret the noise to be the result of the subject applying random forces to its arm. Finally, we define $h = 0.02$.

\begin{figure}
    \begin{minipage}{0.5\textwidth}
        \includegraphics[width=\textwidth]{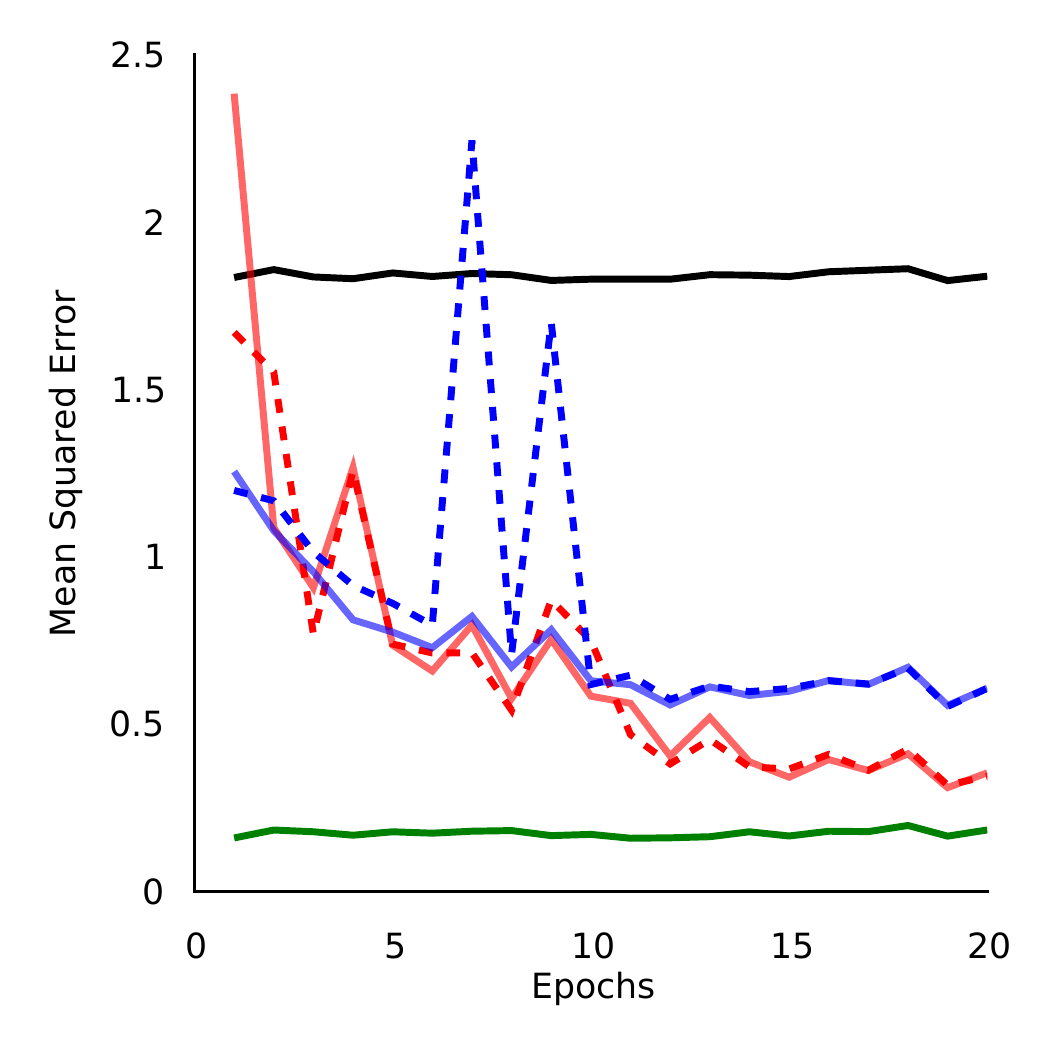}
    \end{minipage}
    \begin{minipage}{0.5\textwidth}
        \includegraphics[width=\textwidth]{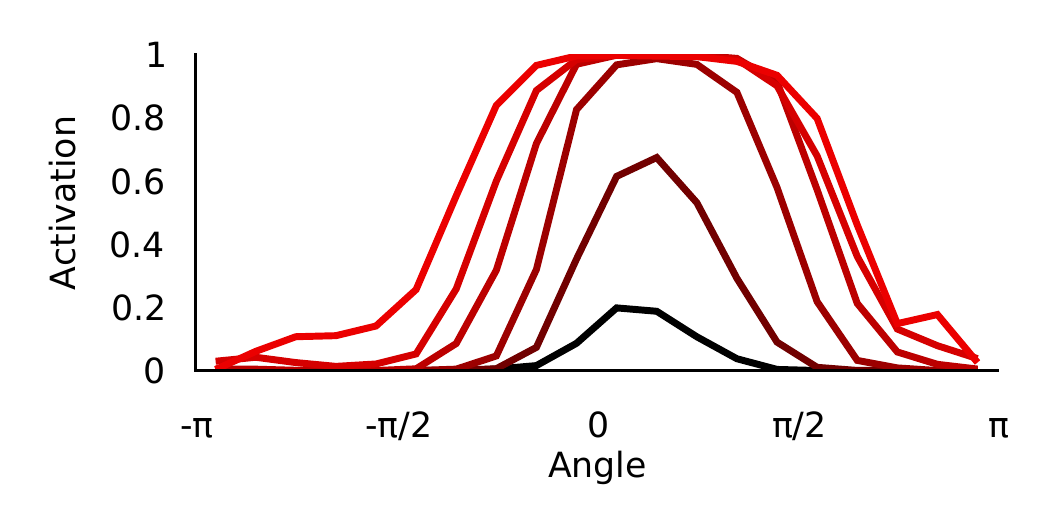}
        \includegraphics[width=\textwidth]{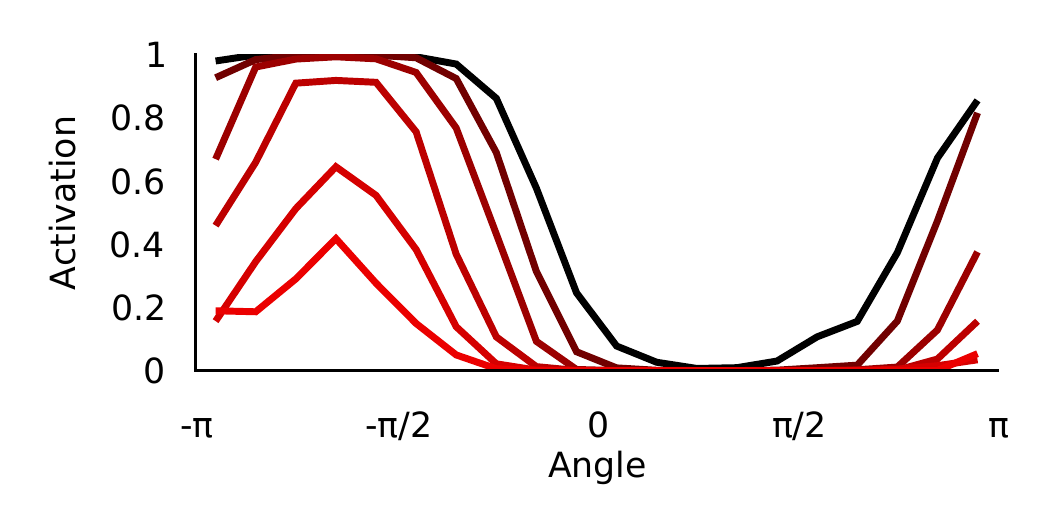}
    \end{minipage}
    \includegraphics[width=\textwidth]{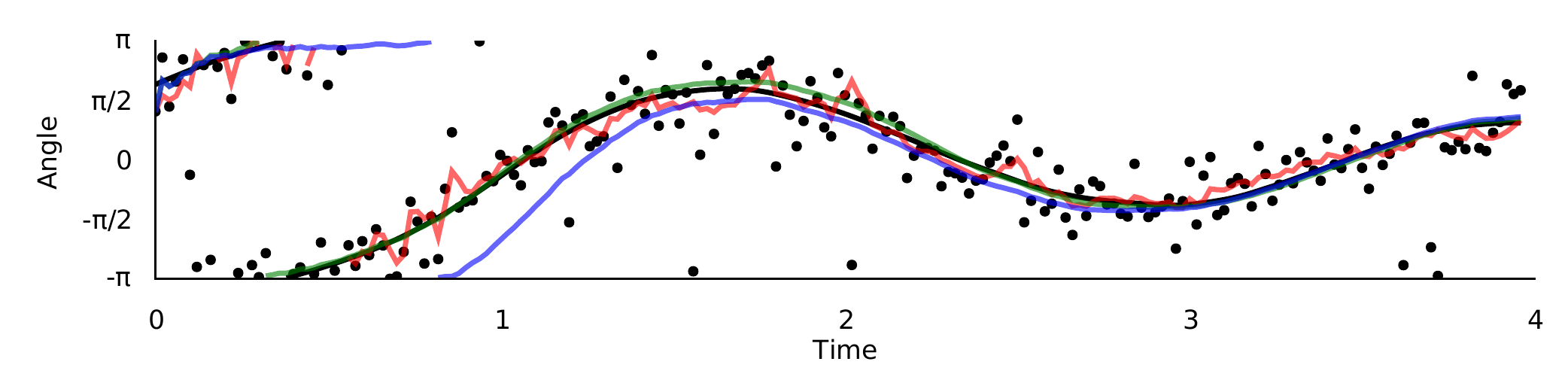}
    \includegraphics[width=\textwidth]{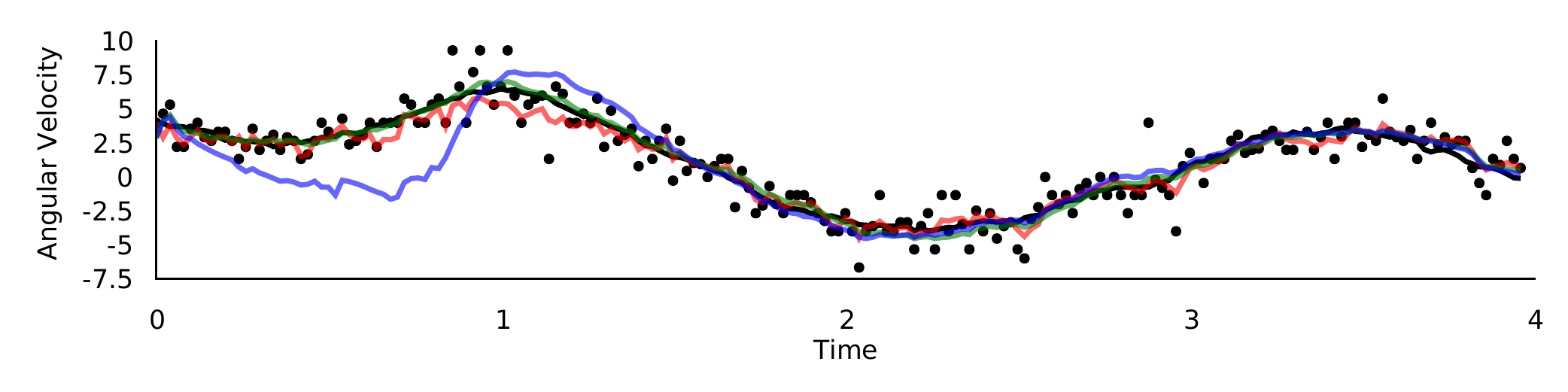}
    \caption{\textbf{Proprioception Training and Simulation}: Here we depict the training of the proposed neural circuits, and simulations with the OT-CD circuit. \emph{Top Right}: The descent of the mean squared error of the approximate beliefs using the same colour scheme in figure \ref{fig:neural-hmm}, where green indicates the approximate EKF beliefs. \emph{Top Right}: Two tuning curves from the hidden layer of the multilayer perceptron. The stimulus angle is plotted on the x-axis, and the stimulus angular velocity is indicated by the colours from red to black, where black indicates an angular velocity of -6, and red indicates a velocity of 6. \emph{Bottom}: Simulation of the angle and angular velocity from the dynamical system. We depict the stimulus (black line), and the dynamic mean of the response posteriors (black dots), the approximate EKF belief density (green line), the approximate KF belief density (blue line), and the learned belief density (red line).}
    \label{fig:pendulum}
\end{figure}

We define the gain of the emission density $p_{N \mid X}$ by $\gamma = h\cdot 100 = 2$, and we define the tuning curves of $p_{N \mid X}$ with two independent sets of tuning curves over the angle and angular velocity, such that half the neurons in the observation population respond to angle, and the other half to angular velocity. Since the angle is periodic, we define the tuning curves over the angle as a set of von Mises tuning curves
\begin{equation*}
    f_i(q) = e^{\kappa \cos (q - q^0_i)},
\end{equation*}
with 10 preferred stimuli $q^0_i$ distributed evenly over the period $[-\pi,\pi]$, and concentration $\kappa = 1/2$. The tuning curves over the angular velocity are again 1-dimensional Gaussian tuning curves as defined in equation \ref{eq:gaussian-tuning-curve} with 10 preferred stimuli distributed evenly over the interval $[-12,12]$, and covariance $\sigma^2 = 4$. The sufficient statistic of the exponential family determined by these tuning curves is $\V s(q,\dot q) = (\cos q,\sin q,\dot q, \dot q^2)$. The matrix $\iprms_N$ is a diagonal block matrix in four quadrants, with the parameters in the upper-left and lower-right quadrants defined by the parameters of the two sets of tuning curves, and with the parameters in the upper-right and lower-left quadrants equal to zero. In total, the neural populations have $d_N = d_Y = d_Z = 20$ neurons, and we set the number of neurons in the hidden layer of the prediction network $\V g$ to $d_H = 500$.

We depict the results of the four simulations in figure \ref{fig:pendulum}. As displayed in the top left panel, the circuit which best approximates the optimal beliefs, by an extremely slim margin over the OT-EF circuit, is the orthogonal circuit trained with the contrastive divergence minimization (dashed red), which achieves $r = 89.7\%$ of the performance of the approximate EKF. In the lower two panels we depict a 4 second simulation from the system. The black dots, green line, and red line indicate the mean of the response posteriors, the approximate EKF, and the learned beliefs, as in the previous section. The blue line indicates the mean of an approximate Kalman filter with linear dynamics given by the small-angle approximation, and which updates its beliefs with the same strategy as the approximate EKF. As can be seen, a straightforward linear model is not sufficient for tracking the nonlinear stimulus.

In the upper right two panels we depict two tuning curves from the hidden layer of the trained multilayer perceptron $\V g$. We plot the two-dimensional tuning curves by plotting the stimulus angle on the x-axis, and indicating the angular velocity with the colour of the line, where black corresponds to -6, and red to 6. As can be seen in these plots, the tuning curve over the angle is a von Mises tuning curve, and the angular velocity is a monotonic function, and the two components interact multiplicatively. Such multiplicative interactions in neural populations are known as gain-fields, and have been widely applied in theory \citep{zipser_back-propagation_1988,sejnowski_spatial_1995,pouget_spatial_1997} and reported in experiment \citep{salinas_gain_2000, hwang_gain-field_2003, paninski_superlinear_2004, herzfeld_encoding_2015}.

\section{Discussion}

In this paper we demonstrated how to define and train a theoretical neural circuit to approximately implement a Bayes filter, and thereby encode accurate beliefs about an unknown, dynamic stimulus. As depicted in figure \ref{fig:coupled-markov-model}, this neural circuit is composed of three neural populations called the observation population, the prediction population, and the filtering population, as well as a neural network called the prediction network. The observation population generates responses to the stimulus, the prediction population encodes predictions of the stimulus, the filtering population encodes beliefs about the stimulus, and the prediction network computes the rates of the prediction population as a function of the rates of the filtering population.

In our work we assume that the parameters of the three neural populations are fixed, and that our goal is exclusively to optimize the parameters of the prediction network. Towards this end, we derived the negative log-likelihood gradient of the parameters of the prediction network given the responses of the observation population. In addition, we developed a novel approach to approximating this gradient based on the theory of exponential families.

We demonstrated our methods in three simulated experiments. In the first experiment we modelled a sequence learning task and neural populations along the ventral stream, in the second we modelled self-localization and neural populations in the visual cortex and hippocampus, and in the third we modelled proprioception optimization, and shoulder joint receptors and neural populations in the cerebellum. In each experiment we demonstrated how our circuit recovered most of the performance of an optimal, or near-optimal filter. Moreover, we found that the hidden layer of the prediction network developed tuning curves which reproduce well-known experimental findings in the literature.

In concluding this paper we discuss three topics. Firstly, we discuss how our work relates to other algorithms for filtering in machine learning and computational neuroscience. Secondly, we provide a deeper analysis of the models developed and trained in the previous section, and the broader relevance of this work for experimental neuroscience. Thirdly and finally, we discuss how the theory we have developed in this paper could be extended to continuous-time spiking networks, and how it could be applied to modelling noise correlations in the brain.

\subsection{Related Computational Work}

To begin, let us consider the relationship between the exponential family gradient (EF) we have introduced in this paper and the contrastive divergence gradient (CD) for approximating the negative log-likelihood gradient of the exponential family harmonium (EFH). As can be seen in the gradient descent panels of figures \ref{fig:neural-hmm}, \ref{fig:attractor}, and \ref{fig:pendulum}, the EF gradient in general perform as well or better than the CD gradient. Although the advantage is relatively slight, the EF gradient is also much easier to compute -- it requires no additional sampling, and no tuning of the number of contrastive divergence steps. As such, when the EFH in question approximately satisfies equation \ref{eq:tuning-curve-sum}, it is arguable that the EF gradient should be applied.

Our method for approximate Bayesian filtering based on an EFH is related to previous work on approximate filtering with restricted Boltzmann machines \citep{sutskever_recurrent_2009,boulanger-lewandowski_modeling_2012}. Although there are many differences in the details, our model can essentially be viewed as a special case of the RNN-RBM model presented in \cite{boulanger-lewandowski_modeling_2012}. However, the predictions and beliefs of Bayesian filtering are not clearly separated in the RNN-RBM, and the updating of the dynamic parameters of the RNN-RBM is rather justified as a ``mean-field'' approximation. In our case, by extending the work on optimal Bayesian inference with probabilistic population codes \citep{ma_bayesian_2006}, we present conditions under which these updates are exact, leading to equations \ref{eq:population-relation}, \ref{eq:population-relation2}, and \ref{eq:encoded-neural-bayes-rule} for optimally implementing Bayes' rule, and allowing us to describe optimal filters \citep{beck_exact_2007,beck_marginalization_2011} as special cases of our general model. In short, the RNN-RBM is arguably overparameterized, as many of the neural circuits it describes suboptimally implement Bayes' rule.

Another importance difference between our model and the RNN-RBM is that we do not apply backpropagation-through-time (BPTT). It has been argued that BPTT is necessary for these models to learn implicit higher-order temporal structure \citep{sutskever_training_2013, makin_recurrent_2016}. In particular, in the experiment of section \ref{sec:proprioception}, if the observation population does not respond to the angular velocity of the arm, then BPTT should be required to infer it. Indeed, in our simulations we have found that training a neural circuit to filter responses to a pendulum fails when the observation population does not respond to the angular velocity. Nevertheless, there are proprioceptors for both position and motion \citep{mccloskey_kinesthetic_1978}, so this fact does not limit our ability to model proprioception with our circuit, and if it were required, we could in principle apply BPTT to infer the missing state variables.

Another model related to our own is the rEFH presented in \cite{makin_learning_2015}, which also depends on an EFH trained with contrastive divergence minimization to approximate a Bayes filter based on the responses of a dynamic Poisson population. Despite these similarities, however, there are important differences between the rEFH and our approach. At every time $k$, the rEFH optimizes the joint density of the rates of the filtering population $Z_k$ and the concatenated vector $(N_k,Z_{k-1})$ of the response of the observation population and the previous rates of the filtering population, whereas in our circuit and the RNN-RBM it is the conditional probability of $N_k$ given $Z_{k-1}$ that is optimized. Moreover, in the rEFH architecture it is this joint density over $Z_k$ and $(N_k,Z_{k-1})$ that is modelled as an EFH \citep{makin_learning_2013}, whereas in our circuit the EFH is over the stimuli $X_k$ and responses $N_k$.

Because the optimization problem is based on the joint density of $Z_k$ and $(N_k,Z_{k-1})$, the structure of the rEFH circuit is more strict, and so cannot, for example, incorporate a multilayer perceptron for computing predictions. Moreover, it is not clear if the rEFH can exactly implement Bayes' rule, or the optimal solutions discussed in sections \ref{sec:closed-form-solutions} and \ref{sec:optimal-filtering-in-neural-circuits}. Nevertheless, as reported in \cite{makin_recurrent_2016}, the rEFH can learn to infer velocities without direct observation, and without using BPTT. Moreover, in our circuit the gradient descent procedure involves computing expectations in the space of the stimulus directly, which is difficult to justify biologically, whereas the rEFH applies Gibbs sampling between two neural populations, which is less problematic. As such, the training procedure of the rEFH is more biologically realistic \cite{makin_learning_2015}, and choosing either our circuit or the rEFH comes down to a trade-off between a flexible prediction network and theoretical exactness on one hand, and a flexible and biologically realistic learning rule on the other.

\subsection{Neuroscience Applications}

The three simulated experiments we presented in this paper were kept theoretically simple so that they could be validated against optimal models, yet they can easily be extended to more complex and realistic experimental designs as required. We consider simulation \ref{sec:colour-sequence-learning} in its current form to constitute a sound experiment, as such a sequence learning experiment could easily be performed with real subjects, allowing hypothetical networks and circuits to be compared against subject performance and recorded neural activity. The self-localization experiment of section \ref{sec:self-localization} could be expanded to 2-dimensional place cells by applying 2-d Gaussian tuning curves, and could incorporate models of spatially periodic grid cells in the entorhinal cortex \citep{moser_place_2008,giocomo_computational_2011} through the application of von Mises tuning curves. Finally, our work can trivially be extended to include control variables, which would allow the proprioception task in section \ref{sec:proprioception} to depend on motor commands and efference copies \citep{thrun_probabilistic_2005,sarkka_bayesian_2013,makin_learning_2015}.

One surprising result of our simulations was the dramatic effect that the choice of population code can have on learning. We considered the naive circuit which uses the same population code across the observation, prediction, and filtering populations, and the orthogonal circuit, which uses the code presented in the supplementary material of \cite{beck_marginalization_2011} for the prediction and filtering populations. In the self-localization and proprioception experiments, the naive circuit performs reasonably well, though not nearly as well as the orthogonal circuit. In the sequence learning experiment, however, the naive circuit fails to even achieve the upper-bound on the error provided by the instantaneous information in the responses. This finding is in line with computational \citep{boulanger-lewandowski_modeling_2012} and experimental \citep{chang_idiosyncratic_2010} evidence which suggests that diverse population codes improves performance in neural circuits.

Moreover, an important feature which distinguishes these two codes is the importance of the sum of the rates of the population. When the sum of the rates of the observation population is constant, then the sum of the rates is also proportional to the precision of the encoded density \citep{ma_bayesian_2006, beck_probabilistic_2007}. In the naive circuit, this implies that the sums of the rates of the prediction and filtering populations are also proportional to the precision of the encoded densities, and since the rates of the three neural populations are always positive, this enforces a trade-off between encoding accurate beliefs and adding too much precision when adjusting the rates of the prediction and filtering populations. In the orthogonal circuit, however, because the rows of the decoding matrix are mutually orthogonal and orthogonal to the vector of ones, , the parameters of the encoded density may be adjusted independently, and the magnitude of the sum of the rates of the filtering population does not influence the encoded beliefs. Evidently, this provides a much better code for learning to implement a Bayes filter.

We have also shown that the hidden layer of the prediction network learns tuning curves over stimuli. In the self-localization experiment (\ref{sec:self-localization}), training the network resulted in sigmoid tuning curves over the unobserved stimuli. Although sigmoid tuning curves are often found in the brain \citep{pouget_spatial_1997,pouget_information_2000}, to the best our knowledge, sigmoid tuning curves for self-location have not been found in the limbic system. Because it is a linear circuit with no hidden activity, the optimal circuit described in equation \ref{eq:optimal-linear-circuit} could avoid this discrepancy. In our experiments, however, we found that a nonlinear network is required for learning stable neural circuits, and that we could not successfully train a neural circuit based on a linear prediction network. Moreover, self-localization in general is a nonlinear problem, for which a linear prediction network would in any case not suffice.

Although it could be the case that there exists an as of yet undiscovered neural population in the limbic system with tuning curves which match those of our learned hidden layer, we rather suspect that the model neural circuit which we tested fails to capture essential features of the self-localization circuitry, and that the sigmoid tuning curves are a result of this. The exact structure and connectivity of recurrent connections in the hippocampus and entorhinal cortex remains a highly active area of research, and we believe that our work can contribute to this research by providing a general framework for exploratory modelling. By matching the observation, prediction, and filtering populations, as well as the prediction network, to hypotheses about neural circuitry, the resulting performance and internal structure of the circuit can serve to validate the hypothesis in question. In our simple case, our experiment emphasizes that a local neural network is insufficient for explaining the activity of hippocampal place cells, and that a more realistic neural circuit must incorporate additional neural circuitry, for example from the entorhinal cortex.

In simulation \ref{sec:proprioception}, training the neural circuit resulted in von Mises tuning curves over the angle, and sigmoid tuning curves over the angular velocity, which interact via multiplication. When the tuning curve over one stimulus interacts multiplicatively with the tuning curve over another stimulus, it is known as a gain-field or gain modulation \citep{salinas_gain_2000}, and gain-fields have been found in many areas of the brain \citep{salinas_gain_2000, hwang_gain-field_2003, paninski_superlinear_2004}. In particular, \cite{herzfeld_encoding_2015} demonstrated that eye position and velocity is encoded by Purkinje cells in the cerebellum with the same gain-field structure as in our arm-localization circuit. Although the stimuli in this experiment and our simulated experiment are different, both respective neural circuits must ultimately predict the motion of parts of the body, and they do so in similar manners.

It is well-known that given data which match population activity for encoding stimuli, gain-fields can arise spontaneously in the hidden layer of multilayer perceptrons \citep{zipser_back-propagation_1988}. At the same time, Gaussian/sigmoid gain-fields have been used to model the neural computation of coordinate transformations in the posterior parietal cortex \citep{sejnowski_spatial_1995,pouget_spatial_1997}, and were found to be especially apt for computing addition over the encoded variables. In our proprioception experiment, although our neural circuit is not performing a coordinate transformation per se, it is learning to add velocity to position at every time, and therefore the emergence of this particular gain-field fits well into the existing theory. What is novel in our work however, is that our neural network is not trained solve a standard regression problem as in \cite{zipser_back-propagation_1988}, but is trained rather as part of a more complex neural circuit for implementing a Bayes filter. As we have demonstrated, gain-fields continue to emerge in this context.

In our simulated experiments we implemented the prediction networks with multilayer perceptrons for performance reasons. Nevertheless, the maximum likelihood approach we have presented in this paper can be applied to any parameterized network, and it is entirely possible to apply our method to optimize the parameters of more biologically plausible prediction networks and thereby validate more realistic neural circuit models. To reiterate, our theory is not dependent on multilayer perceptrons, but rather only populations of Poisson-spiking neurons and probabilistic population codes, which are well-established for explaining the activity of populations of neurons \citep{dayan_theoretical_2005,pouget_probabilistic_2013}.

\subsection{Future Directions}

In concluding our paper we discuss two ways in which we hope to extend and apply our work in the future. In particular, we discuss how to model a continuous-time, spiking neural circuit with our methods, and how our model might be applied to understanding noise correlations in the brain.

Although we described the neural circuit for linear Bayesian filtering presented in \cite{beck_marginalization_2011} as a special case of our model, the circuit in \cite{beck_marginalization_2011} is both a spiking and continuous-time circuit, which ours is not. Nevertheless, extending our neural circuit to be spiking and continuous-time is relatively straightforward. On one hand, as shown in \cite{beck_marginalization_2011}, generating spikes from the rates of the prediction and filtering populations and using them as the exclusive basis for neural communication ultimately results in little loss of information. On the other hand, a parameterized network with the form of \ref{eq:optimal-linear-circuit} can in principle be optimized by our method, and the linear transformations could be made nonlinear for stability and more expressive power. In unpublished work we have done exactly this, and the initial results are promising. Nevertheless, there are details and pitfalls specific to the training of such a continuous-time circuit, which are beyond the scope of this paper.

Understanding noise correlations in neural populations is a major research area in neuroscience, as they represent a breakdown of the simple understanding of neurons as independent Poisson processes, and affect the efficacy of neural coding \citep{averbeck_neural_2006}. The only source of noise in our neural circuit is in the observation population, which is indeed a population of independent Poisson neurons, and the rates of the prediction and filtering populations are only random by virtue of being functions of the observation population. This might suggest that our theoretical neural circuit cannot model noise correlations, however recent research has shown that many patterns of noise correlation in the brain can be explained by correlations resulting from downstream responses to sensory noise \citep{kanitscheider_origin_2015}. In initial simulations we have indeed found that the prediction and filtering populations in our neural circuits exhibit significant noise correlations, and in the future we hope to use our neural circuit model to explore the extent to which noise correlations in dynamic neural populations can be explained as the result of sensory noise.

\section*{Acknowledgements}

This work was partially funded by the DFG Priority Program 1527, Autonomous Learning. The author would like to thank Nihat Ay, Guido Montufar, Keyan Zahedi, and Anna Erzberger, for their comments, advice, and support.

\bibliographystyle{apalike}
\bibliography{./article.bbl}

\end{document}